\documentclass[pdflatex,sn-vancouver-num]{sn-jnl}% Vancouver Numbered Reference Style
\usepackage{graphicx}%
\usepackage{multirow}%
\usepackage{amsmath,amssymb,amsfonts}%
\usepackage{amsthm}%
\usepackage{mathrsfs}%
\usepackage[title]{appendix}%
\usepackage{xcolor}%
\usepackage{textcomp}%
\usepackage{manyfoot}%
\usepackage{booktabs}%
\usepackage{algorithm}%
\usepackage{algorithmicx}%
\usepackage{algpseudocode}%
\usepackage{listings}%

\usepackage{todonotes}
\usepackage{caption}
\usepackage{subcaption}
\usepackage{arydshln} % hdashline
\theoremstyle{thmstyleone}%
\theoremstyle{thmstyletwo}%

\theoremstyle{thmstylethree}%

\usepackage{fancyhdr}
\begin{document}

\title[Article Title]{Explainable time-series forecasting with sampling-free SHAP for Transformers}


\author*[1]{\fnm{Matthias} \sur{Hertel}}\email{matthias.hertel@kit.edu}
\author[1]{\fnm{Sebastian} \sur{P\"utz}}\email{sebastian.puetz@kit.edu}
\author[1]{\fnm{Ralf} \sur{Mikut}}\email{ralf.mikut@kit.edu}
\author[1]{\fnm{Veit} \sur{Hagenmeyer}}\email{veit.hagenmeyer@kit.edu}
\author[1]{\fnm{Benjamin} \sur{Sch\"afer}}\email{benjamin.schaefer@kit.edu}

\affil[1]{\orgdiv{Institute for Automation and Applied Informatics (IAI)}, \orgname{Karlsruhe Institute of Technology (KIT)}, \orgaddress{\street{Hermann-von-Helmholtz-Platz 1}, \city{76344~Eggenstein-Leopoldshafen}, \country{Germany}}}

\abstract{
Time-series forecasts are essential for planning and decision-making in many domains. Explainability is key to building user trust and meeting transparency requirements. Shapley Additive Explanations (SHAP) is a popular explainable AI framework, but it lacks efficient implementations for time series and often assumes feature independence when sampling counterfactuals. We introduce \mbox{SHAPformer}, an accurate, fast and sampling-free explainable time-series forecasting model based on the Transformer architecture. It leverages attention manipulation to make predictions based on feature subsets. SHAPformer generates explanations in under one second, several orders of magnitude faster than the SHAP Permutation Explainer. On synthetic data with ground truth explanations, SHAPformer provides explanations that are true to the data. Applied to real-world electrical load data, it achieves competitive predictive performance and delivers meaningful local and global insights, such as identifying the past load as the key predictor and revealing a distinct model behavior during the Christmas period.
}




\maketitle
%\newpage

\begin{figure}[hbt]
    \centering
    \makebox[\linewidth][c]{
        \includegraphics[width=1.2\linewidth]{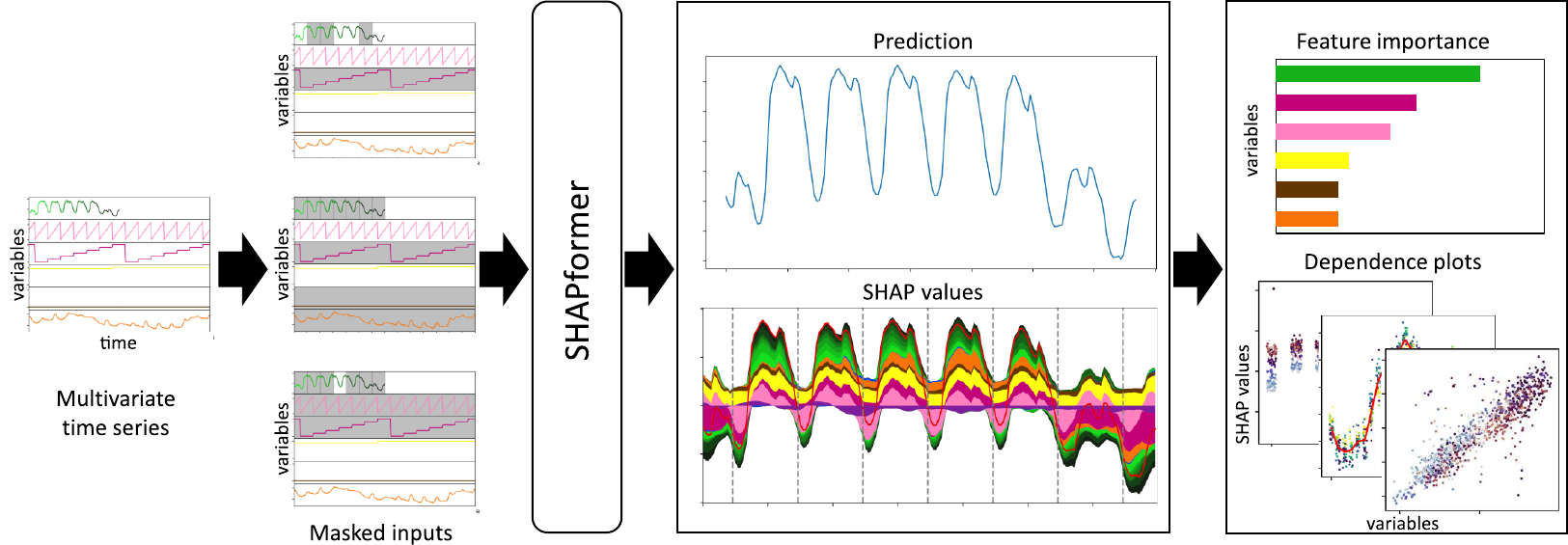}
    }
    \caption{Overview of the proposed method. The model receives the past target and covariate time-series as input. The features are grouped as indicated by colors. SHAPformer makes predictions based on masked inputs. SHAP values are derived from the marginal feature contributions, defined as the difference of the prediction with and without a feature group. Multiple local explanations of forecasts get combined into global feature importance values and feature dependence plots.}
    \label{fig:graphical-abstract}
\end{figure}

\pagestyle{fancy}

\section{Introduction}\label{sec:introduction}
% (767 words)

% \begin{itemize}
%     \item Why forecasting (in energy)
%     \item Why explainability
%     \item Research gap
%     \item Contribution: SHAPformer
%     \item Comparison methods + experimental setup
% \end{itemize}

Time-series forecasting plays a critical role in numerous domains, including logistics, retail, finance, healthcare, business services, traffic management, and energy systems \cite{petropoulos_forecasting_2022}.
In the energy sector, forecasts are becoming increasingly important due to the ongoing transition to renewable energy sources \cite{sweeney_future_2020}, which is essential to mitigate the impacts of anthropogenic climate change \cite{lee_ipcc_2023}.
The inherent variability of renewable generation makes accurate forecasts of electrical load and generation crucial to ensure real-time balance between supply and demand \cite{machowski_jan_and_lubosny_zbigniew_and_bialek_janusz_w_and_bumby_james_r_power_2020}. Moreover, forecasts enable early detection of critical grid conditions and help prevent equipment overloads through demand-side response and redispatch measures \cite{haben_review_2021}.

Across domains, modern forecasting models increasingly rely on complex deep learning architectures \cite{lim_time-series_2021, su_systematic_2025}.
These models often involve a large number of parameters, which makes their internal processes opaque to humans.
This has driven the development of Explainable Artificial Intelligence (XAI) methods, which aim to improve the understanding of how machine learning models operate \cite{ali_explainable_2023}.
Human interpretability is valuable for several reasons.
First, it can enhance user trust and facilitate the adoption of deep learning in real-world applications \cite{das_opportunities_2020}.
Second, explanations help users assess the reliability of predictions and decide how to act on them \cite{panigutti_role_2023}.
Third, they assist developers in debugging and refining models—for example, by revealing spurious patterns or ``Clever Hans" effects \cite{lapuschkin_unmasking_2019}.
Finally, XAI supports compliance with emerging regulations that demand transparency in AI systems. The European Union’s AI Act, for instance, mandates transparency for applications that involve humans or critical infrastructure such as energy systems \cite{european_commission_artificial_2024}. Although not always required, XAI can support human oversight and improve transparency \cite{panigutti_role_2023}, and may justify future regulatory standards for transparency in AI systems.

While comparably simple models, including linear regressors and generalized additive models, are inherently interpretable, more complex and powerful models often require post-hoc explanation methods to make their predictions understandable \cite{adadi_peeking_2018}.
Post-hoc explanation methods are often perturbation-based, which means that they systematically modify the model inputs and observe the corresponding changes in the model output.
%This is necessary because complex models are not directly interpretable due to the large amount of model parameters, unlike simple models such as linear regressors and generalized additive models.
A widely used post-hoc XAI framework is Shapley Additive Explanations (SHAP) \cite{lundberg_unified_2017}.
It is the most popular XAI framework in the energy sector \cite{machlev_explainable_2022} and has been used to explain electrical load forecasts \cite{baur_explainability_2024} produced by models such as long short-term memories \cite{wu_explainable_2022, henriksen_electrical_2022}, multilayer perceptrons \cite{bolstad_day-ahead_2022} and tree-based algorithms \cite{li_power_2022, lee_shap_2023, han_impact_2014, wang_interpretable_2025}.
Based on cooperative game theory, SHAP estimates the contribution of each feature to the prediction and satisfies the efficiency property, which means that the contributions sum up to the prediction \cite{lundberg_unified_2017}.
SHAP is more informative than mere feature attribution methods, as it gives not only feature importance but also the direction (positive or negative) and magnitude of a feature's impact on the prediction.
SHAP supports local explanations of individual predictions, and global explanations such as feature importance scores and dependence plots, which are aggregations of many local explanations \cite{lundberg_local_2020}.
Other post-hoc XAI methods, including LIME \cite{ribeiro_why_2016}, Grad-CAM \cite{selvaraju_grad-cam_2017} and Layer-wise Relevance Propagation (LRP) \cite{bach_pixel-wise_2015, achtibat_attnlrp_2024}, only highlight important inputs without indicating how they affect the prediction, or do not fulfill the efficiency property.

Several algorithms for the estimation of SHAP values have been developed \cite{chen_algorithms_2023}, which quantify the marginal contribution of each feature by evaluating a model based on subsets of features.
These estimations typically follow one of two strategies: (a) sampling absent features from a marginal or conditional distribution, or (b) replacing absent features with a predefined baseline value. Both approaches have limitations. Sampling is computationally intensive and can produce unrealistic inputs that fall outside the data distribution, leading to off-the-manifold evaluations \cite{frye_shapley_2021}. Baseline substitution requires careful selection of the baseline and only explains deviations from that specific reference point.
TimeSHAP \cite{bento_timeshap_2021} is a SHAP version for time series that estimates SHAP values for variables and time steps based on replacing data points with the mean value.
WindowSHAP \cite{nayebi_windowshap_2023} estimates SHAP values for time windows, where absent windows are sampled or replaced.

For Transformer models \cite{vaswani_attention_2017}, it is common to use attention weights as proxies for feature importance, visualizing them to highlight influential inputs \cite{clark_what_2019, hertel_evaluation_2022}. However, this practice remains controversial, as the interpretability of attention mechanisms is debated \cite{jain_attention_2019, wiegreffe_attention_2019}.
The Temporal Fusion Transformer (TFT) \cite{lim_temporal_2021} computes feature importance values with a dedicated feature selection layer, but it gives no information about how the features influence the prediction, and it lacks the efficiency property of SHAP.
An efficient implementation of SHAP for time-series Transformers is currently missing \cite{baur_explainability_2024}.

In this context, we present a new algorithm to estimate SHAP values for time-series forecasting models based on attention manipulation \cite{deiseroth_atman_2023}.
Our contributions are as follows:
\begin{enumerate}
    \item We introduce SHAPformer, a Transformer-based forecasting model that enables the efficient calculation of exact SHAP values by grouping features and applying attention manipulation \cite{deiseroth_atman_2023} to exclude absent feature groups.
    This approach eliminates the need for sampling or setting baseline values. An overview of the method is shown in Figure \ref{fig:graphical-abstract}.
    \item We validate SHAPformer’s explanations on a synthetic load forecasting dataset with known ground truth explanations, demonstrating its ability to accurately capture the underlying patterns.
    %\item We apply SHAPformer to real-world electrical load data from TransnetBW \cite{wiese_open_2019}, the transmission system operator (TSO) in the German state of Baden-Württemberg. In this setting, SHAPformer is several orders of magnitude faster than the widely used SHAP Permutation Explainer and provides meaningful local and global insights into the model’s behavior.
    \item We showcase the benefits of SHAPformer applied to empirical load data from the transmission system operator (TSO) TransnetBW \cite{wiese_open_2019}. SHAPformer is several orders of magnitude faster than the widely used SHAP Permutation Explainer and provides meaningful local and global insights.
\end{enumerate}

\section{Results}
% (1634 words)

%An overview on SHAPformer is given in Figure \ref{fig:graphical-abstract} and the implementation details are presented in Section \ref{sec:methods}.
% Comparison methods
% We compare SHAPformer with multiple established explainable forecasting methods.
% First, we use the Temporal Fusion Transformer (TFT) \cite{lim_temporal_2021}, which is a Transformer-based model using feature selection networks to compute feature importance.
% Second, we train a Transformer forecasting model and use the SHAP Permutation Explainer to create local and global explanations.
% Third, we use the same Transformer model and implement a Custom Masker for the SHAP Permutation Explainer, which treats daily load curves and exogenous time series as features instead of sampling every input value independently, therefore reducing the number of features and increasing the runtime compared to the default Permutation Explainer.
% Finally, a persistence baseline, which predicts the observed pattern from the week before, is used as a comparison model that is inherently interpretable.
% All approaches are evaluated on two datasets: a synthetic dataset with daily, weekly and yearly seasonality and dependencies on external factors (as presented in Section \ref{sec:methods}), and an empirical dataset with electrical load and weather data from the German state Baden-Württemberg.

We evaluate SHAPformer on two datasets:
\begin{itemize}
    \item Synthetic dataset: We generate a synthetic dataset exhibiting daily, weekly, and annual seasonality with dependencies on exogenous covariates (holidays and a multiplier), for which ground truth explanations are available The details of the data generation are presented in Section \ref{sec:synthetic-dataset}. 100,000 examples are used for training and 10,000 each for validation and testing. 
    \item TransnetBW dataset: This dataset contains hourly electrical load measurements from the German TSO TransnetBW for the years 2015--2019 \cite{wiese_open_2019} and weather data from the Copernicus ERA5 reanalysis model \cite{copernicus_climate_change_service_climate_2020}. The last six months are used for testing and the six months before for validation.
\end{itemize}

SHAPformer is compared against a standard Transformer model explained by two SHAP algorithms, the Permutation Explainer and a Custom Masker, as well as against state-of-the-art forecasting models (detailed in Section \ref{sec:methods}).
All models have one week context length and forecast the 168 hourly values for the following week.
Forecast accuracy, training time, and inference time for all methods are summarized in~Table~\ref{tab:results}.
For the real-world dataset, average results from five repeated runs with different model initializations are reported.
Additional metrics and the standard deviations for the five runs are reported in Appendix \ref{app:detailed-results-transnet}.

\begin{table}[hbt]
    \caption{Test root mean squared error (RMSE), training time and inference time on the synthetic dataset and real dataset. Inference time is the time needed to create one explanation for the Permutation Explainer, Custom Masker and SHAPformer, and to create one forecast for all others.}
    \label{tab:results}
    \setlength{\tabcolsep}{6pt}
    \renewcommand{\arraystretch}{1.15}
    \begin{tabular}{lccccccc}
        \toprule
              Approach & SHAP & \multicolumn{2}{c}{Forecast error (RMSE)} & \multicolumn{2}{c}{Training time [h]} & \multicolumn{2}{c}{Inference time [s]} \\
         & values & Synthetic & Real [MW] & Synthetic  & Real            & Synthetic & Real             \\
        \midrule
        Persistence baseline    & no & 0.152 & 652.3 & \phantom{0}- & - & \phantom{00}- & \phantom{00}- \\
        Linear Regression       & no & 0.149 & 553.7 & \phantom{0}0.00 & 0.00 & \phantom{000}0.00 & \phantom{00}0.00\\
        XGBoost                 & no & 0.119 & 387.0 & \phantom{0}0.01 & 0.00 & \phantom{000}0.00 & \phantom{00}0.00 \\
        Temporal Fusion Transformer & no & 0.059 & 390.8 & \phantom{0}5.20 & 1.84 & \phantom{000}0.01 & \phantom{00}0.03 \\
        %\hdashline
        Transformer             & no & 0.059 & 263.1 & \phantom{0}0.90 & 0.26 & \phantom{000}0.01 & \phantom{00}0.01 \\
        + Permutation Explainer & approximate & " & " & \phantom{0}" & " & 1124.16 & 484.34 \\
        + Custom Masker         & approximate & " & " & \phantom{0}" & " & \phantom{000}7.84 & \phantom{00}3.54 \\
        SHAPformer              & exact & 0.060 & 265.9 & 10.55 & 3.46 & \phantom{00}21.90 & \phantom{00}0.60 \\
        \bottomrule
    \end{tabular}
\end{table}

\subsection{Fast calculation of exact SHAP values while maintaining forecast quality}

As SHAPformer is based on the Transformer architecture, we compare it to a Transformer forecasting model.
Explanations for the Transformer are generated with two SHAP algorithms:
(1) the Permutation Explainer, which samples absent features from background data under the assumption of feature independence; and (2) a Custom Masker, which groups features into seven load-related groups (one for each of the past seven days) and one group for each exogenous variable. SHAP values are then computed per group, instead of per feature, by jointly sampling all features within a group.

SHAPformer generates explanations more than $50\times$ faster than the Permutation Explainer on the synthetic data, and more than $800\times$ faster on the real-world data.
%between $2\times$ and over $300\times$ faster than the other SHAP algorithms.
Remarkably, this is achieved although SHAPformer calculates exact SHAP values based on all feature group subsets, whereas the Permutation Explainer and Custom Masker approximate SHAP values based on ten random feature permutations.
The Custom Masker achieves a 136--143$\times$ speedup by reducing the number of feature subsets through grouping.
In addition, SHAPformer uses attention manipulation to drop absent feature groups and thereby eliminates the need for sampling and repeated model evaluations, so that it becomes feasible to run it with all feature group subsets.
However, SHAPformer’s inference speed advantage comes at the cost of increased training time, ranging from two to ten times longer than that required by TFT or the standard Transformer.
The inference times of the Transformer and SHAPformer are larger on the synthetic data than on the real-world data, because the models are larger (see Appendix \ref{sec:hyperparameters} for the hyperparameters) and the synthetic dataset contains one covariate more than the real-world dataset.

TFT produces explanations with a single model evaluation, making it the fastest Transformer-based explanation method. However, it is not directly comparable to the SHAP-based approaches, as it provides only feature importance scores but does not provide information on how individual features influence the predictions.

% The forecast quality of SHAPformer is better or comparable to the other models.
% All evaluated models are better than a persistence baseline predicting the value from the week before.
% The Transformer is the best forecasting model on both datasets, closely followed by SHAPformer, which has an approximately 1\% higher forecast error.
% %SHAPformer has almost the same quality as the Transformer on the synthetic data, and a 2.4\% higher error on the empirical data.
% On the empirical data, both the Transformer and SHAPformer give more accurate forecasts than the Linear Regression, XGBoost \cite{chen_xgboost_2016} and the Temporal Fusion Transformer.
% % On the synthetic dataset, SHAPformer has a 2.9\% higher MSE than TFT and the Transformer, and on the TransnetBW data, the MSE of SHAPformer is 2.4\% higher than that of the second-best model.
% % All three Transformer-based models are much better than the persistence baseline on both datasets.
% The inference speedup gained by SHAPformer comes at an increased training cost, 
% with two to ten times the training time of TFT and the Transformer.
% %with a ten-fold increase in training time compared to the Transformer and double the training time of the Temporal Fusion Transformer.
% %The training time of SHAPformer compared to the Transformer model is twelve times higher on both datasets, and is roughly double the training time of the Temporal Fusion Transformer.

SHAPformer's forecast accuracy is comparable to or better than the compared models. All evaluated models outperform the persistence baseline, which simply predicts the value from one week earlier. The Transformer achieves the lowest forecast error on both datasets, with SHAPformer close behind, with a forecast error that is only about a 1\% higher. On the real-world dataset, both the Transformer and SHAPformer outperform Linear Regression, XGBoost \cite{chen_xgboost_2016}, and TFT.
The larger forecast error of TFT on the real-world data stems from (a) a larger variance across experiment repetitions (but SHAPformer and the Transformer outperform TFT in all runs), and (b) from large forecast errors before Christmas, where TFT's causal attention prohibits the model from using the holiday feature of the later days when predicting the first days in the forecast horizon.

\subsection{Successful validation of the explanations on synthetic data}
\label{sec:results-synthetic}

Having demonstrated SHAPformer’s forecast accuracy and fast inference, we validate its explanations using synthetic data with known ground truth explanations. The synthetic time series incorporate daily, weekly, and annual seasonality, along with dependencies on covariates. SHAP is used on the data generation process in order to compute ground truth explanations. Details of the dataset and ground truth computation are provided in Section \ref{sec:synthetic-dataset}.

\begin{figure}
    \hspace{-1.5cm}A Feature importance\\
    \makebox[\linewidth][c]{
    \hspace{1.5cm}
    \begin{subfigure}{\linewidth}
        \includegraphics[width=0.8\textwidth]{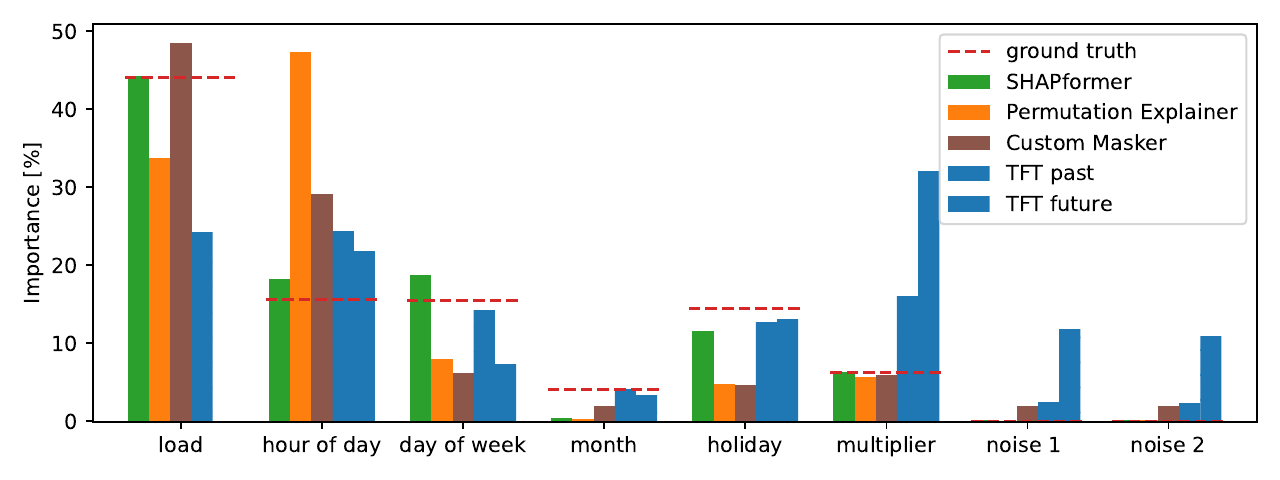}
    \end{subfigure}
    }

    \hspace{-1.5cm}B SHAPformer dependence plots\\
    \makebox[\linewidth][c]{
        \includegraphics[width=1.15\linewidth]{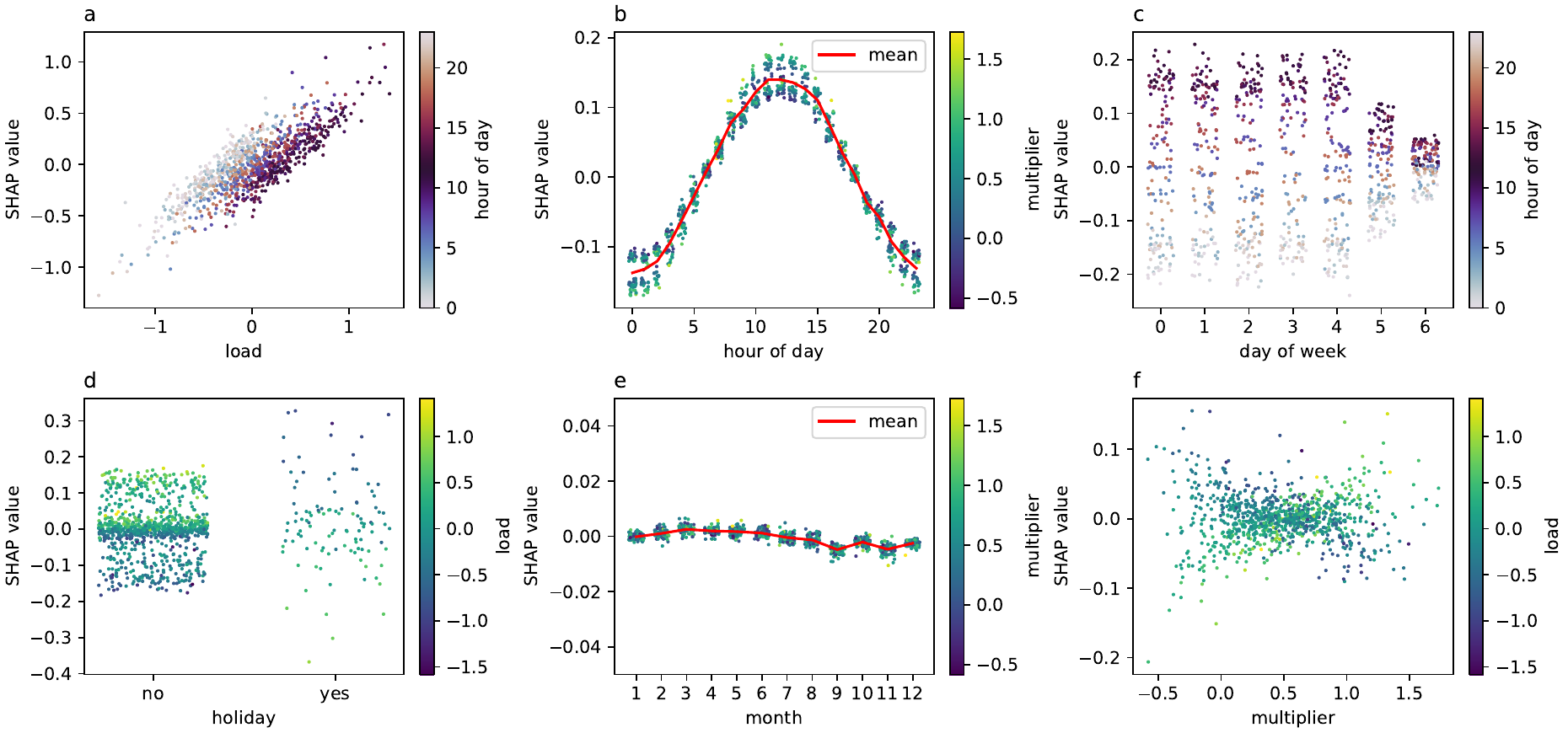}
    }

    % \hspace{-3cm}B SHAPformer dependence plots\\
    % \makebox[\linewidth][c]{
    % \begin{subfigure}{0.5\linewidth}
    %     \includegraphics[width=\textwidth]{figures/synthetic/load_hourofday.pdf}
    % \end{subfigure}
    % \begin{subfigure}{0.5\linewidth}
    %     \includegraphics[width=\textwidth]{figures/synthetic/hod_temperature.pdf}
    % \end{subfigure}
    % \begin{subfigure}{0.5\linewidth}
    %     \includegraphics[width=\textwidth]{figures/synthetic/dow_hourofday.pdf}
    % \end{subfigure}
    % }
    % \makebox[\linewidth][c]{
    % \begin{subfigure}{0.5\linewidth}
    %     \includegraphics[width=\textwidth]{figures/synthetic/holiday_load.pdf}
    % \end{subfigure}
    % \begin{subfigure}{0.5\linewidth}
    %     \includegraphics[width=\textwidth]{figures/synthetic/month_temperature.pdf}
    % \end{subfigure}
    % \begin{subfigure}{0.5\linewidth}
    %     \includegraphics[width=\textwidth]{figures/synthetic/temperature_load.pdf}
    % \end{subfigure}
    % }

    \hspace{-1.5cm}C Ground truth dependence plots\\
    \makebox[\linewidth][c]{
        \includegraphics[width=1.15\linewidth]{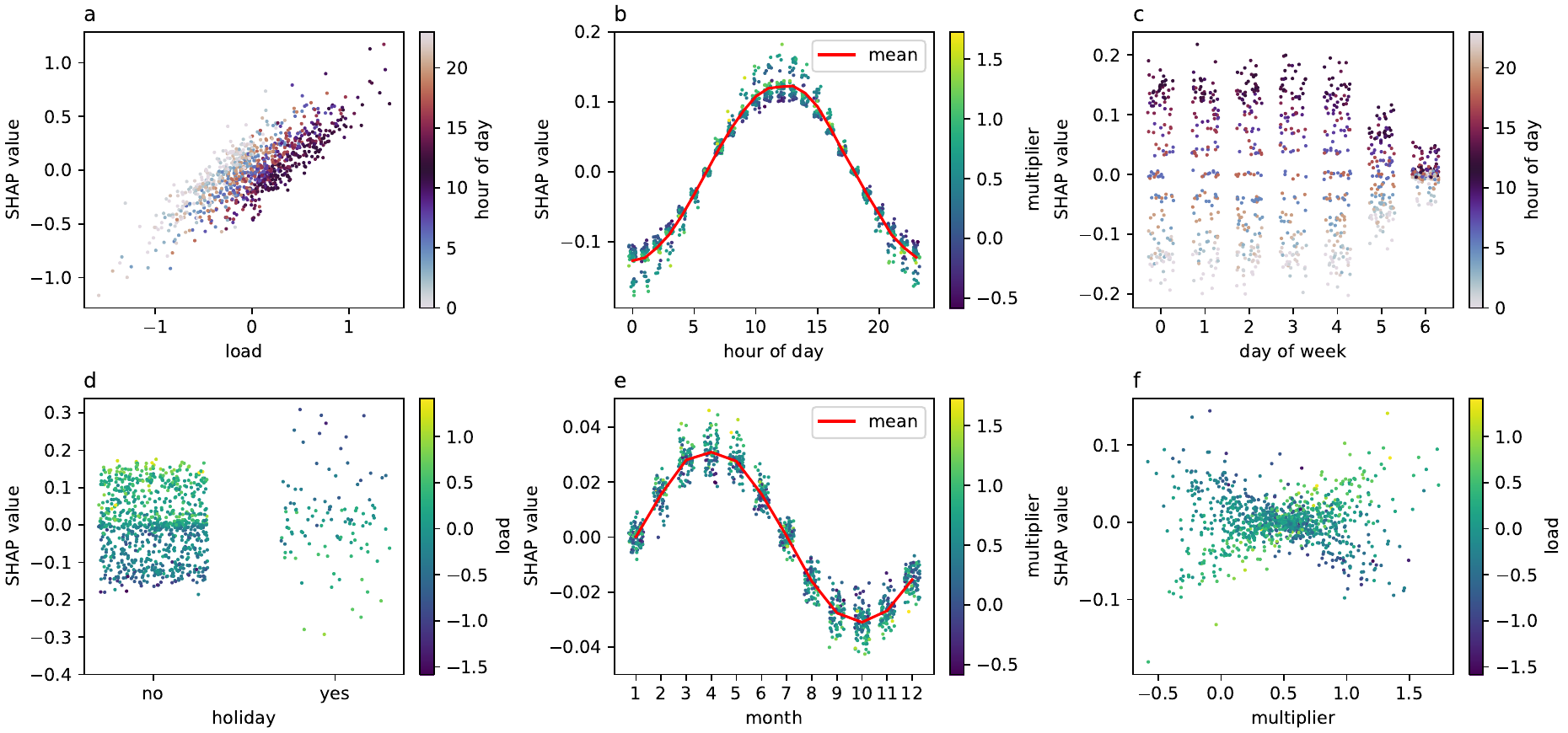}
    }
    
    % \hspace{-3cm}C Ground truth dependence plots\\
    % \makebox[\linewidth][c]{
    % \begin{subfigure}{0.5\linewidth}
    %     \includegraphics[width=\textwidth]{figures/ground_truth/load_hourofday.pdf}
    % \end{subfigure}
    % \begin{subfigure}{0.5\linewidth}
    %     \includegraphics[width=\textwidth]{figures/ground_truth/hod_temperature.pdf}
    % \end{subfigure}
    % \begin{subfigure}{0.5\linewidth}
    %     \includegraphics[width=\textwidth]{figures/ground_truth/dow_hourofday.pdf}
    % \end{subfigure}
    % }
    % \makebox[\linewidth][c]{
    % \begin{subfigure}{0.5\linewidth}
    %     \includegraphics[width=\textwidth]{figures/ground_truth/holiday_load.pdf}
    % \end{subfigure}
    % \begin{subfigure}{0.5\linewidth}
    %     \includegraphics[width=\textwidth]{figures/ground_truth/month_temperature.pdf}
    % \end{subfigure}
    % \begin{subfigure}{0.5\linewidth}
    %     \includegraphics[width=\textwidth]{figures/ground_truth/temperature_load.pdf}
    % \end{subfigure}
    % }

    \caption{Global explanations on the synthetic test data. A: SHAPformer approximates ground truth feature importances well (TFT: Temporal Fusion Transformer). B and C: dependence on the six most important features. The strongest interacting feature is shown in color (interactions mean that the SHAP value depends on the interacting variable). For discrete variables, noise was added in the x-direction for visibility reasons.}
    \label{fig:global-explanations-synthetic}
\end{figure}

SHAPformer's global feature importance is close to the ground truth, except that it underestimates the importance of the month feature (Figure \ref{fig:global-explanations-synthetic}A). In contrast, the Permutation Explainer and Custom Masker deviate from the ground truth, particularly for key features such as the past target (called "load" as for the real-world data), hour of day, and day of week. TFT is more difficult to interpret, as it produces separate sets of feature importance values for past and future features. Nonetheless, it is observable that neither of the sets resembles the ground truth, nor does a linear combination of the two.
Notably, both SHAPformer and the Permutation Explainer effectively filter out irrelevant features such as the two noise covariates, while the other methods assign them importance values greater than zero, although the noise covariates are not related to the target variable.

In addition to feature importance, we compare SHAPformer’s learned feature dependencies (Figure \ref{fig:global-explanations-synthetic}B) with the ground truth (Figure \ref{fig:global-explanations-synthetic}C), where the strongest interacting variable is indicated by the color of the dots. Overall, SHAPformer’s feature dependencies align well with the ground truth, with the only exception being the underestimation of the month feature importance.
In panel (a) of Figure \ref{fig:global-explanations-synthetic}B-C, the previous week’s load has a linear effect on the prediction.
The same past load results in a higher SHAP value at night (bright dots) than during the day (dark dots), because a normal daytime load is already exceptionally high for the night.
In panel (b), the dependence on the hour of day forms a half-sine wave, which is exactly the daily pattern used in the synthetic data generation. This interacts with the multiplier: negative multipliers (dark blue dots) decrease the effect of the hour of day, whereas positive values (light green to yellow) increase the effect.
Panel (c) reflects the multiplicative shrinking effect used in the data generation to model weekends (days five and six), with SHAP values closer to zero than on weekdays -- negative at night, positive during the day. Panel (d) shows that holidays shrink the predicted load by increasing negative past loads and reducing positive ones, while non-holidays exhibit a reversed, but less pronounced, effect. In panel (e), the month feature has a sinusoidal effect in the ground truth, which SHAPformer partially captures, though with much reduced amplitude. Finally, panel (f) shows an X-shaped pattern for the multiplier feature: higher multipliers amplify positive previous loads and reduce negative ones, whereas lower multipliers have the opposite effect.

Taken together, the fact that SHAPformer's feature importance and feature dependencies resembles the ground truth confirms that it captures the true dependencies present in the data.
Next to the global explanations, SHAPformers local explanations are also consistent with the ground truth, which is shown in Appendix~\ref{app:local-explanations}.
The dependence plots and local explanations created with the Permutation Explainer and Custom Masker are presented in Appendix~\ref{app:shap-explanations} for comparison.

\subsection{Insights into load forecasts on real-world TSO data}

\begin{figure}
    \hspace{-1.5cm}A Feature importance\\
    \makebox[\linewidth][c]{
    \hspace{1.5cm}
    \begin{subfigure}{\linewidth}
        \includegraphics[width=0.8\textwidth]{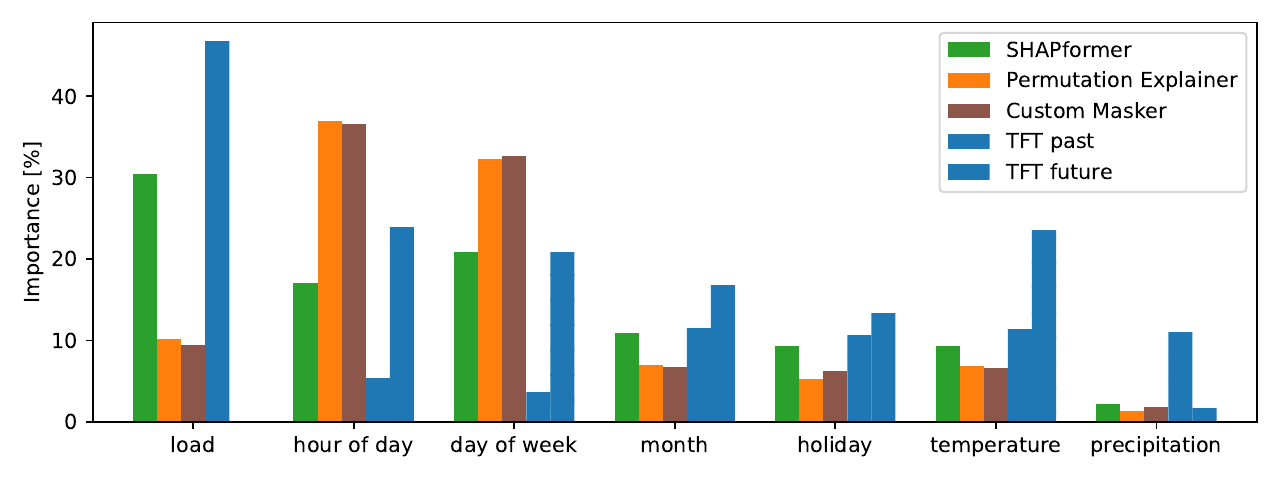}
    \end{subfigure}
    }

    \hspace{-1.5cm}B SHAPformer dependence plots\\
    \makebox[\linewidth][c]{
        \includegraphics[width=1.15\linewidth]{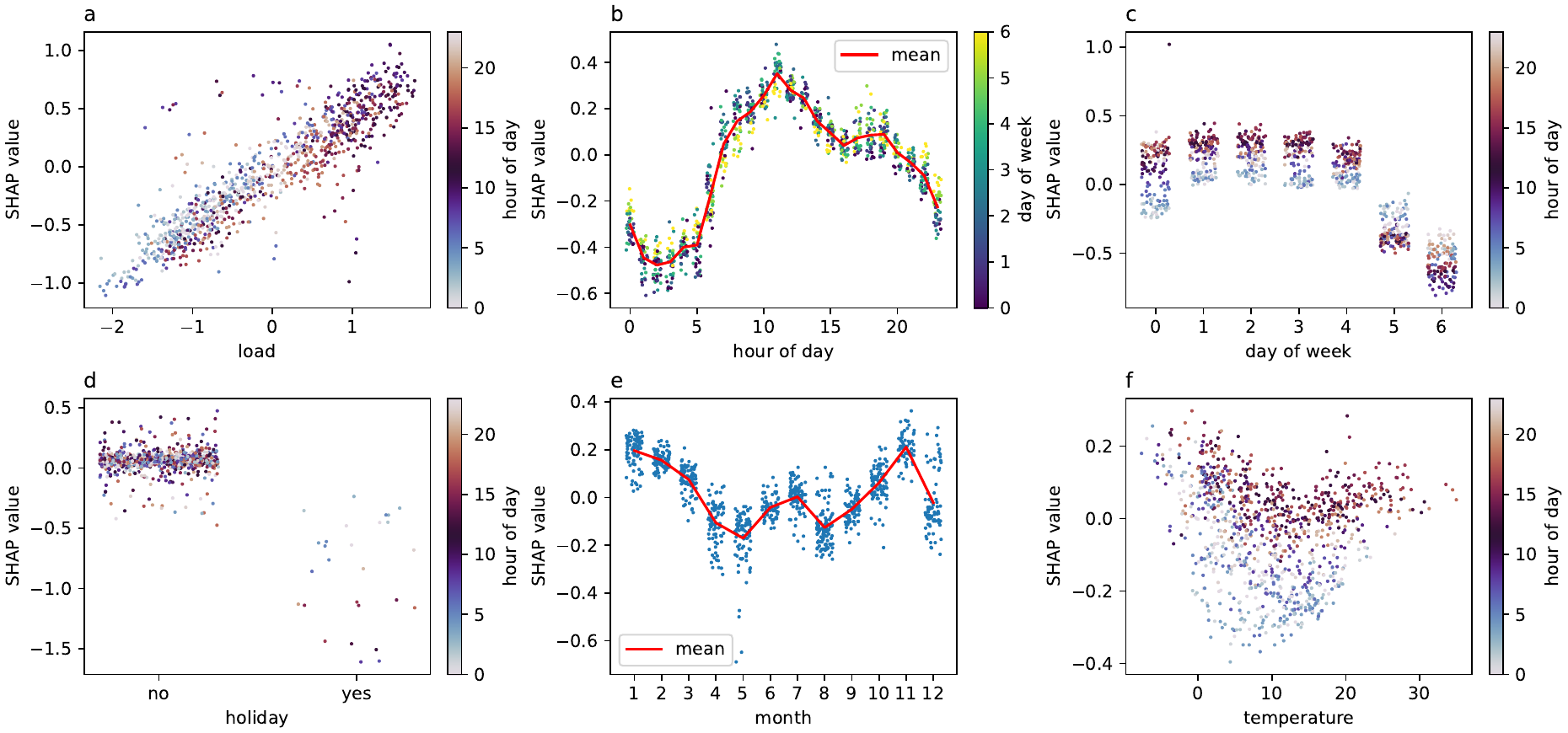}
    }

    % \hspace{-3cm}B SHAPformer dependence plots\\
    % \makebox[\linewidth][c]{
    % \begin{subfigure}{0.5\linewidth}
    %     \includegraphics[width=\textwidth]{figures/transnet/load_hourofday.pdf}
    % \end{subfigure}
    % \begin{subfigure}{0.5\linewidth}
    %     \includegraphics[width=\textwidth]{figures/transnet/hod_dayofweek.pdf}
    % \end{subfigure}
    % \begin{subfigure}{0.5\linewidth}
    %     \includegraphics[width=\textwidth]{figures/transnet/dow_hourofday.pdf}
    % \end{subfigure}
    % }
    % \makebox[\linewidth][c]{
    % \begin{subfigure}{0.5\linewidth}
    %     \includegraphics[width=\textwidth]{figures/transnet/holiday_hourofday.pdf}
    % \end{subfigure}
    % \begin{subfigure}{0.5\linewidth}
    %     \includegraphics[width=\textwidth]{figures/transnet/month.pdf}
    % \end{subfigure}
    % \begin{subfigure}{0.5\linewidth}
    %     \includegraphics[width=\textwidth]{figures/transnet/temperature_hourofday.pdf}
    % \end{subfigure}
    % }
    
    \caption{Global explanations on real-world load data from TransnetBW. A: Feature importance scores by SHAPformer, a Transformer model explained using the Permutation Explainer and the Custom Masker, and Temporal Fusion Transformer (TFT). B: Dependence plots from SHAPformer, with feature values on the x-axis and corresponding SHAP values on the y-axis. The strongest interacting variable is indicated by color. For discrete variables, noise was added in the x-direction for visibility reasons.}
    \label{fig:global-explanations-transnet}
\end{figure}

With SHAPformer validated on synthetic data, we examine its global explanations for the last 12 months of the real-world dataset (Figure \ref{fig:global-explanations-transnet}). In this case, no ground truth is available, so the quality of the explanations is assessed based on domain knowledge.
%, which can be particularly difficult due to interacting variables.

% The load of the previous week is the most important feature, followed by the day of week and hour of day features. The month, temperature and holiday features are less important and show similar importance among each other, whereas precipitation has almost no importance.
% The Permutation Explainer and the Custom Masker agree almost perfectly in their explanations of the Transformer.
% Here, the hour of day and day of week features are the most important, and the load is much less important than for SHAPformer.
% %, followed by the other features, with precipitation being the least important.
% %For TFT, an overall interpretation is difficult, because the model returns two sets of feature importance (one for features available in the past and one for features available in the future), whose relation to each other is unknown. Load is the most important past feature, whereas day of week and holidays are the most important future features. Multiple past features are considered less important than the past precipitation, but precipitation is the least important future feature.
% For TFT, an overall interpretation is difficult because the model returns two sets of feature importances, whose relation to each other is unknown. The load is the most important past feature, and the hour of day, temperature and day of week are the most important future features.

SHAPformer identifies the past load as the most important predictor, followed by the day of the week and the hour of the day. Features such as month, temperature, and holidays contribute less and show similar levels of importance, while precipitation has almost no importance. The explanations of the Permutation Explainer and Custom Masker are similar to each other but differ from SHAPformer's explanations, emphasizing hour of day and day of week, and assigning much lower importance to the past load than SHAPformer. For TFT, the interpretation is not straightforward, as it outputs two distinct sets of feature importance -- one for past inputs and one for future covariates -- with an unclear relationship between them. In TFT’s case, the most important past feature is the load, while hour of day, temperature, and day of week are the most important future features.

Figure \ref{fig:global-explanations-transnet}B displays SHAPformer’s learned feature dependencies. SHAP values are expressed in terms of standardized electrical load, where a value of 1.0 corresponds to one standard deviation (approximately 1550 MW). In panel~(a), a linear relationship between the previous week's load and the forecasted load is observed. Notably, SHAP values are higher at night (bright dots) for the same load value -- likely because a load considered low during the day can be unusually high at night.
Panel (b) shows the typical daily load pattern with lower values at night and two peaks around noon and in the evening. On Sundays (yellow dots), SHAP values tend to be closer to zero. Panel (c) illustrates that the predicted load is lower on Saturdays and Sundays, especially during the day (dark dots). Holidays, shown in panel (d), consistently reduce the predicted load, although the effect is weaker at night (bright dots). In panel (e), the month feature reflects a seasonal pattern with higher loads in winter -- except in December, likely due to reduced industrial activity during the Christmas period. Finally, panel (f) shows that temperature increases the predicted load on cold days, particularly when daytime temperatures (bright dots) are below 15\,°C or nighttime temperatures (dark dots) are below 0\,°C.

% Figure \ref{fig:local-explanations-transnet} shows two local explanations from SHAPformer in December. In both plots, the hour of day feature decreases the prediction at night and shows two daily peaks, and the day of week feature decreases the prediction on the weekend, as in the dependence plots.
% The month has a positive effect on the left-hand side, despite the fact that the week is in December -- this is likely because the past week is in November. On the right-hand example, which is later in December, the month effect is strongly negative.
% The holiday effect is positive during the week. This indicates that the model has learned to predict a lower load in December, except for weeks without holidays.
% On the left-hand side, the first predicted days are cold, resulting in an increased prediction. The predicted days on the right-hand side are all mild, resulting in a lower prediction.
% The past load increases the prediction on the weekdays in the left-hand example, but decrease it on days 3 to 7 on the right-hand side -- especially due to the last two days of the previous week which show a low average load.
% % Why?

\begin{figure}
    \hspace{-1.5cm}A Data\\    
    \makebox[\linewidth][c]{
    %\hspace{1.5pt}
    \begin{subfigure}{0.8\linewidth}
        \centering Feature groups \& target \\
        \includegraphics[width=\textwidth]{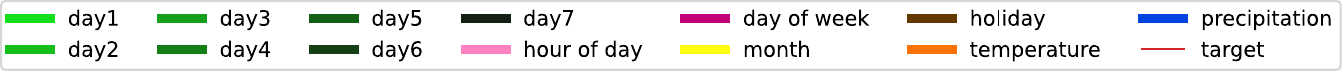}
    \end{subfigure}
    }
    \vspace{-1cm}
    
    \makebox[\linewidth][c]{
    \begin{subfigure}{0.6\linewidth}
        \includegraphics[width=\textwidth]{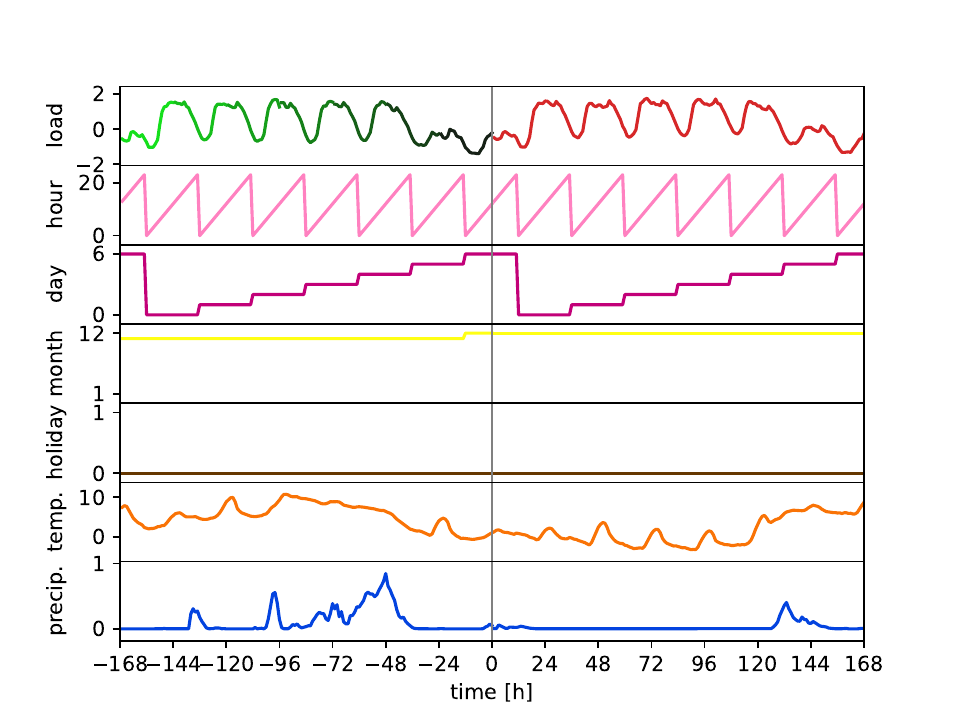}
    \end{subfigure}
    \begin{subfigure}{0.6\linewidth}
        \includegraphics[width=\textwidth]{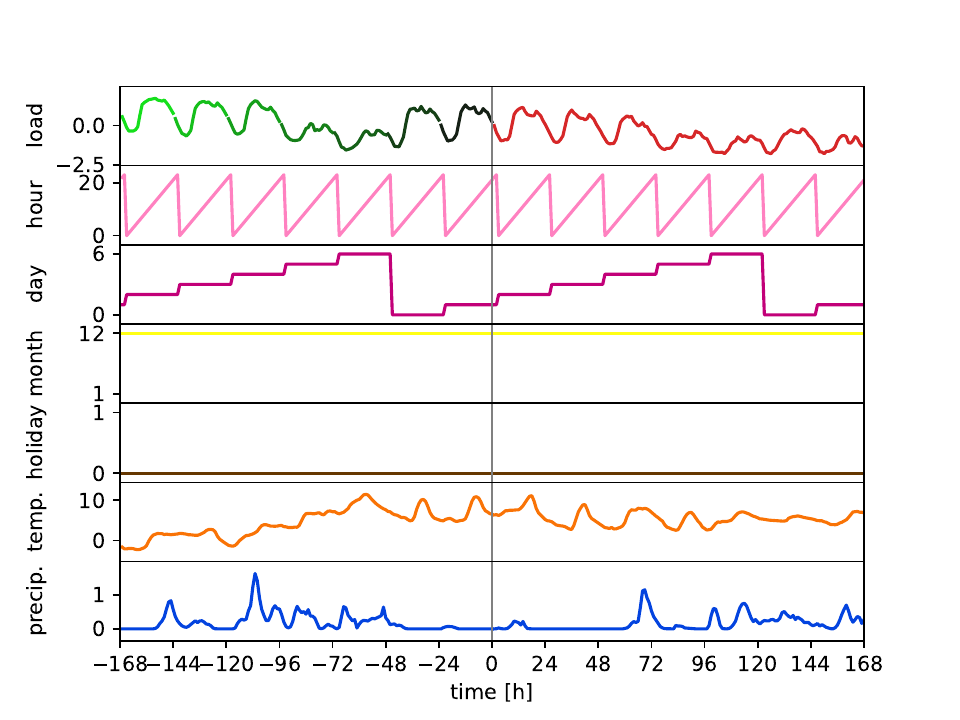}
    \end{subfigure}
    }

    \hspace{-1.5cm}B SHAPformer\\
    \makebox[\linewidth][c]{
    \begin{subfigure}{0.6\linewidth}
        \includegraphics[width=\textwidth]{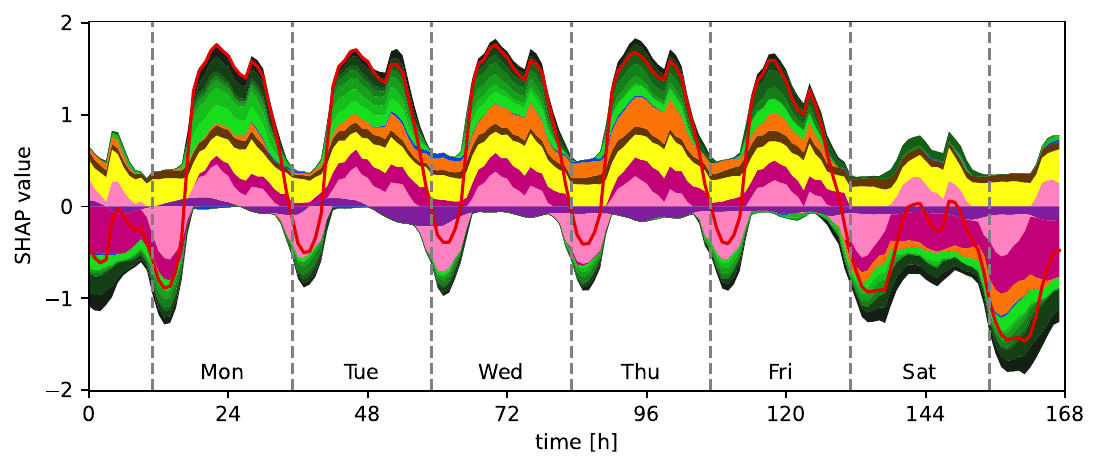}
    \end{subfigure}
    \begin{subfigure}{0.6\linewidth}
        \includegraphics[width=\textwidth]{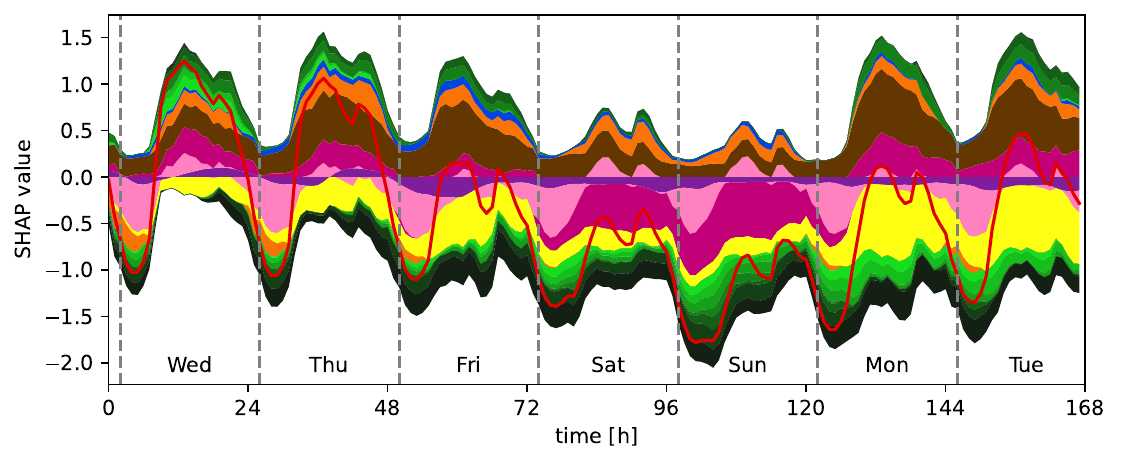}
    \end{subfigure}
    }

    \caption{Local explanations of examples from the real-world load data of TransnetBW. The start of the forecast horizon on the left-hand side is December 1, 2019, 13:00 and on the right-hand side December 17, 2019, 22:00. Panel B shows the prediction as a red line.}
    \label{fig:local-explanations-transnet}
\end{figure}

Figure \ref{fig:local-explanations-transnet} presents two local SHAPformer explanations for forecasts in December. In both cases, the hour of day feature lowers the predictions at night and shows two daily peaks -- mirroring the patterns seen in the dependence plots. The day of week feature reduces the predictions on weekends, as expected. Interestingly, in the left-hand example, the month feature has a positive effect despite being in December -- likely because the previous week falls in November, which is represented in the month feature of the encoder input. In contrast, the right-hand example, which is later in December, shows a strongly negative month effect.
The holiday feature has a positive impact during weekdays, suggesting that the model associates December with lower loads in general due to the long holiday season, except during weeks without holidays.
In the left-hand example, the cold temperature in the early forecast days raises the prediction, while the warmer temperature in the last two days decreases the prediction. In the right-hand example, the temperature is more constant and has a consistently positive effect on the prediction. The effect of the past load also differs in the two examples: in the left-hand plot, the past load increases the prediction on weekdays, whereas in the right-hand plot, it decreases the prediction from days 3 to 7 -- especially influenced by the last two input days which show a lower average load than the rest.

Overall, the observed dependence patterns align well with domain knowledge, reinforcing confidence in SHAPformer’s predictions. The December pattern -- where a lower load is treated as the default and non-holidays as exceptions -- is unexpected, but can be plausibly explained by the extended holiday period during that time.

% Figure \ref{fig:local-explanations-transnet} shows the local explanations of three examples of the TransnetBW test data.
% SHAPformer is highly impacted by the load of the week before, whereas the Transformer model looks more on the calendar feature -- an observation that could also be made in the global explanations.
% The first example is a week in December before Christmas. Despite the December usually having a negative impact on the prediction by SHAPformer, it is positive in this example, probably due to the fact that the month in the impact week is November.
% The second example is a week in December including Christmas eve as the last day (which is no public holiday). SHAPformer shows an interesting pattern: The effect of the December is negative, but the fact that there are no holidays cancels out the December effect -- opposite to the normal holiday behavior, that the holiday feature has only a little effect on non-holidays but decreases the load on holidays.
% The third example is a week with the Christmas holidays in the beginning. Here, the holiday feature has a negative effect not only on the holidays, but also on the days around them, as many people take vacation during these days.
% The cold outside temperature contributes to a higher load.

\section{Discussion}
% (908 words)

%Pro & Con of SHAPformer
We introduced SHAPformer, an accurate and efficient approach for explainable time-series forecasting. SHAPformer enables the generation of SHAP explanations for Transformer-based models several orders of magnitude faster than the established SHAP Permutation Explainer. This significant speedup results from two key mechanisms: grouping input features to reduce the number of coalitions, and estimating marginal contributions without sampling by manipulating the attention weights. Both mechanisms contribute to reduced runtime. The Custom Masker, which applies feature grouping but not attention manipulation, is already faster than the Permutation Explainer. However, only SHAPformer achieves inference times below one second per explanation on the real-world data.

SHAPformer’s fast inference comes at the cost of increased training time. Training with masked inputs requires the model to learn a more complex structure across varying feature subsets, and since it only sees partial inputs in each iteration, the effective amount of training data per epoch is reduced. Nevertheless, for applications where many forecasts need to be explained, the reduced inference time justifies the additional training effort.

The explanation runtime of SHAPformer increases exponentially with the number of feature groups, because the model is evaluated for all $2^N$ subsets of $N$ feature groups.
In our setup with seven input days and six or seven covariates (depending on the dataset), this full enumeration is computationally feasible and desirable as it yields the most accurate SHAP estimates.
However, in scenarios with more feature groups, such as longer input sequences or additional covariates, this exhaustive approach may become impractical.
In such cases, SHAPformer can be adapted to evaluate only a subset of the feature coalitions, enabling a runtime that scales linearly with $N$.
% SHAPformer is fast enough to allow the evaluation of all feature subsets. With a large number of exogenous variables or multiple weeks input, this might no longer be feasible, since the number of subsets grows exponentially with the number of feature groups. In a scenario with many input days, it might be desirable to drop the calculation of SHAP values for each day and instead treat the complete past load as a covariate.
% In the future, SHAPformer can be adapted to run only on a subset of the $2^N$ coalitions, and SHAP values be estimated from the predictions, similar to the Permutation Explainer. In this case, the runtime would only grow linearly with the number of feature groups.

% validation on synthetic data
SHAPformer has been successfully validated on synthetic data, as it closely reproduces the ground truth in terms of feature importance, dependence patterns, and feature interactions. It effectively filters out irrelevant noise features and captures key relationships, including daily patterns, multiplicative effects of holidays and weekends, and temperature interactions. The only exception is the month feature, whose importance SHAPformer appears to underestimate, likely because its influence in the data generation process is small and comparable to the noise added to the target time series (see Section \ref{sec:synthetic-dataset}).

% We make our synthetic dataset and corresponding ground truth explanations publicly available to support the evaluation of future explanation methods. However, it's important to note that our ground truth relies on specific assumptions: namely, that explanations should resemble SHAP values and respect known feature dependencies (e.g., inferring hour of day from transitions in the day of week).
% Under different assumptions, such as explaining a model that has learned other dependencies, alternative explanations may also be valid.
We publish the synthetic dataset and the ground truth explanations, so that they can be used in the future for the evaluation of XAI methods\footnote{The synthetic dataset with ground truth explanations is available on GitHub: \url{https://github.com/KIT-IAI/SHAPformer}}. However, note that this ground truth is created based on multiple assumptions, such as that the ground truth should resemble SHAP values and it should respect dependencies between features (e.g. the hour of day can be inferred from the jumps in the day of week feature). Other explanations might be valid under different assumptions, especially when the goal is to create explanations that are true to a model that has learned different dependencies.
%e.g. when the goal is to compute explanations which are true to a trained model instead of true to the data.
We view this evaluation approach as complementary to existing explanation metrics \cite{nauta_anecdotal_2023} and believe that it can be extended to other data modalities in the future.

Existing SHAP algorithms are either true to the model or true to the data \cite{chen_true_2020}.
Conditional sampling tends to produce explanations that are true to the data, while off-the-manifold sampling generates explanations that better reflect the model’s internal behavior \cite{frye_shapley_2021}.
We argue that SHAPformer achieves both: it is true to the model and true to the data. This is made possible by its sampling-free design and training strategy, which builds robustness to absent features.
In the presence of correlated features, SHAPformer learns to predict accurately using any of the correlated inputs, distributing the predictive contribution among them.
This contrasts with models trained on the full feature set, which may rely on arbitrary feature combinations, leading to explanations that misrepresent underlying data relationships.
SHAPformer's strong alignment with the ground truth explanations on synthetic data
%, generated to reflect the dependencies in the data, 
supports this claim.

%In our experiments, we noted a major difference in the importance of the hour of day feature importance between SHAPformer and the Transformer model. Probably, the Transformer relies on the hour of day feature as it encodes the order of the input vectors, and the prediction changes strongly when this feature is altered.
%However, an explanation being true to the data would not rate the hour of day feature as important, because it can be inferred from various other features, such as the day of the week (which shows the day boundaries) or the temperature (which shows a daily pattern).
%Indeed, SHAPformer distributes the importance among these correlated features and is therefore more true to the data.

% The sampling-free estimation of marginal contributions with SHAPformer might also result in better explanations, as it does not rely on generating forecasts on unrealistic out-of-distribution samples, nor does one have to deal with sampling of highly correlated inputs, which occur naturally in time series.
% %On the downside, SHAPformer needs more time to train because it sees only a subset of inputs with each training example.
% Compared to TFT, SHAPformer has the advantage that it returns SHAP values, which can be used to create local and global explanations, and they can be directly compared between past and future features.

% insights into learned model
Applying SHAPformer to real-world load data provides valuable insights into how different features influence the model's predictions. The most important predictor is the load from the previous week, followed by the day of the week and the hour of the day. Other features -- such as month, temperature, and holidays -- also contribute, while precipitation has minimal impact. This contrasts with the Transformer model explained by the Permutation Explainer or Custom Masker, which emphasizes temporal features more heavily and downplays the role of the past load. This difference highlights how global explanations can help differentiate between models with similar forecast accuracy but different learned dependencies. In our view, a forecasting model can reasonably rely on both calendar-based inputs or lagged load values to estimate typical load patterns. SHAP values are instrumental in uncovering which of these cues the model relies on. Given that SHAPformer aligns more closely with the ground truth explanations on synthetic data, we are confident that its explanations on real-world data are likewise more faithful to the underlying data relationships than those from the comparison methods.
In the future, the efficient SHAP calculation of SHAPformer can be adopted for other models than Transformers, by training them on feature subsets, which allows to run them on feature subsets at inference time.
This can be achieved by using feature dropout instead of attention manipulation.
Training models in this way not only facilitates explainability but may also improve robustness to known anomalies and missing values. By dropping out these values with attention manipulation or equivalent mechanisms, such models can adapt more effectively to real-world data imperfections.
%One option to drop out a variable is setting its value to zero. This could be misleading in some cases, since the zero often has a meaning (e.g., after standardization, zero is the variable's mean value). A value different from all observed values could be used (e.g., a large negative value). Alternatively, a binary flag can be introduced for each variable, indicating whether this variable is dropped. The model can then learn to distinguish between a normal zero value and zero representing a dropped-out variable. Then, SHAP values can be computed efficiently without sampling. A drawback is the potentially higher training time, as observed with SHAPformer.
%Attention manipulation for missing values
% Models trained on feature subsets can be used in the future to increase robustness to known anomalies and missing values, by manipulating the attention so that these values are not used.
%In the same way that attention manipulation is used in SHAPformer to create predictions based on feature subsets, it can be used to drop out time steps or variables that should not be used by the model due to anomalies or measurement errors.

We release SHAPformer as a Python package to make it available to researchers and practitioners\footnote{The code for SHAPformer is available on GitHub, with example files demonstrating its use: \url{https://github.com/KIT-IAI/SHAPformer}}.
This will facilitate its application and evaluation across a range of domains beyond electrical load forecasting. Given its successful validation on synthetic data, we are confident that SHAPformer will generalize well to other use cases.

%SHAPformer will also be applied and tested to other domains, beyond empirical electricity load data. With the generic synthetic data set, we are confident that it will generalize well.

%\clearpage

\section{Methods}
\label{sec:methods}

This section introduces SHAP and two algorithms to estimate SHAP values, the Permutation Explainer and the Custom Masker, in Section \ref{sec:shap}.
Next, our method SHAPformer is described in Section \ref{sec:shapformer}.
Then, the synthetic dataset and ground truth generation processes are described in Section \ref{sec:synthetic-dataset}.
Finally, the compared forecasting models are described in Section \ref{sec:baselines}.

\subsection{Shapley Additive Explanations (SHAP)}
\label{sec:shap}

\subsubsection*{Shapley values}

Shapley values are a concept from game theory that is used to compute a player's contribution to the outcome of a cooperative game \cite{shapley_lloyd_s_value_1953}.
The player's contribution is defined as how much the game outcome changes on average when the player enters the game, considering all permutations of players entering the game.
The Shapley value definition is the unique solution for the computation of players' contributions that fulfill the efficiency axiom (the sum of the contributions adds up to the outcome), the symmetry axiom (two identical players receive equal contributions), the dummy axiom (a player that does not contribute to any coalition gets a contribution of zero) and the additivity axiom (the contribution to the sum of two games equals the sum of the contributions of the two games) \cite{lundberg_unified_2017, molnar_interpreting_2023}.

\subsubsection*{Explaining machine learning models with SHAP}

The concept of Shapley values is used in machine learning to compute the contribution of features to the prediction of a model \cite{lundberg_unified_2017, molnar_interpreting_2023}.
For this, the features are the players and the model prediction is the game outcome.
The exact Shapley values usually cannot be computed for computational reasons and because it is not possible to integrate over the unknown distributions of absent features.
Therefore, the Shapley values are estimated in practice and are then called SHAP values, in order to distinguish them from the exact Shapley values.
The naive algorithm to compute SHAP values works as follows:
For a set of features $\{v_1, ..., v_n\}$, consider all permutations $p_j$, $1 \leq j \leq n!$.
For a permutation $p_j$, go through all features in the order of the permutation.
For a feature $v_i$, let $S$ be the features that precede $v_i$ in $p_j$.
Then, the marginal contribution of $v_i$ is computed as:
\begin{equation}
    \phi(v_i, p_j) = f(S \cup \{v_i\}) - f(S)
\end{equation}
where $f(X)$ is the model prediction using a set of features $X$.
The marginal contribution is the effect on the model prediction when adding $v_i$ to the feature set.
The SHAP value for $v_i$ is defined as the average marginal contribution of $v_i$ in all permutations.
It can be computed more efficiently by iterating over all $2^n$ feature subsets instead of all $n!$ feature permutations:
\begin{equation}
    \operatorname{SHAP}(v_i) = \sum_{S \subseteq V \setminus \{v_i\}} \dfrac{(n - 1 - |S|)! \cdot |S|!}{n!} \cdot (f(S \cup \{v_i\}) - f(S))
\end{equation}
The first part of the formula is a weight, defined by the fraction of permutations in which $v_i$ gets added to $S$, and the second part is the marginal contribution of $v_i$, defined as the difference of the model prediction with and without $v_i$.
Note that to compute all feature contributions, it is necessary to compute the model predictions for all $2^n$ feature subsets $S \subseteq V$ (also called coalitions) \cite{lundberg_unified_2017, molnar_interpreting_2023}.

Usually, a machine learning model trained on a set of features requires all features for inference, which means that it does not allow for predictions based on feature subsets.
Therefore, a common approach to estimate predictions based on feature subsets is to marginalize out the effect of the absent features via a Monte Carlo integration -- that is, by sampling their values $k$ times from a background dataset (e.g., the training data).
Consequently, the model is called $k$ times for each of the feature subsets.

\subsubsection*{Permutation Explainer}

The Permutation Explainer
%\footnote{See \url{https://shap.readthedocs.io/en/latest/generated/shap.PermutationExplainer.html}.}
is a more efficient way of estimating SHAP values by calculating marginal contributions only for a subset of the coalitions \cite{molnar_interpreting_2023}.
It samples $K$ random feature permutations. For each permutation $p_j$, it starts with the empty feature set and iteratively adds features in the order of the permutation. The marginal contribution of each feature is computed as the prediction with that feature minus the prediction without that feature, i.e. $f(S \cup \{v_i\}) - f(S)$ as above, where $S$ is the set of features preceding $v_i$ in $p_j$.
As in the naive algorithm, predictions based on feature subsets are estimated by a Monte Carlo integration over the absent features (i.e. by sampling their values from background data). Each permutation is used twice, where features are added once in the forward and once in the backward direction.
The SHAP value estimate is then computed as the average marginal contribution of a feature over all sampled permutations.

The Permutation Explainer is commonly used in practice, but it has two drawbacks:
\begin{enumerate}
    \item It is costly to compute for large feature sets. In our forecasting setup on the TransnetBW dataset with $168 \times 7$ past features and $168 \times 6$ future features (2\,184 features in total), ten feature permutations (each run forward and backward), and 100 samples to estimate distributions of absent features, $2\,184 \times 10 \times 2 \times 100 = 4\,368\,000$ model calls are needed to compute a single local explanation of a forecast.
    The runtimes reported in Table \ref{tab:results} are measured with ten permutations.
    As this is not feasible to evaluate for many samples, we instead reduce the number of permutations to two to generate the local and global explanations shown in the figures.
    \item When feature dependencies are not considered during Monte Carlo sampling, unrealistic samples are created, which are out of distribution of the training data and can potentially lead to arbitrary model behavior. For example, when calendar features are sampled independently, it can happen that Monday midnight is followed by Friday noon. Other examples of unrealistic counterfactuals are 30°C in winter, quick jumps in the electrical load or temperature values, high temperatures or loads at night, and many more.
\end{enumerate}

Note that many algorithms exist that estimate SHAP values \cite{chen_algorithms_2023} and not all of them require sampling.
For text and images, usually no sampling is required. Instead, patches of images are blurred or replaced by unicolored pixels. Words are replaced by ”...” or removed entirely. Similarly, it is possible to replace time series features by a baseline value of zero, their negative values or the values in reversed order, as is done in perturbation methods \cite{schlegel_deep_2023}.
However, choosing a baseline value is not straightforward, and the chosen baseline affects the explanations. After standardization of the continuous time-series features, a value of zero represents the time series' mean, so applying a zero baseline would only explain the difference to the mean.
When a non-holiday is chosen as the baseline for the holiday feature, non-holidays will be attributed SHAP values of zero, as the feature value and the baseline value are the same. For the day-of-week feature, no meaningful sequence of weekdays can be chosen as a baseline, as every sequence appears equally often, whereas constant values (e.g. seven consecutive Mondays or 168 times midnight) are unrealistic.

\subsubsection*{Custom Masker}

We develop a Custom Masker to mitigate the sampling and efficiency issues of the Permutation Explainer.
Given an example for which an explanation is to be computed, and a feature coalition (i.e. subset of the features), the Custom Masker defines how the features absent in the coalition get replaced in order to generate alternative samples which are used in the Monte Carlo estimation.

First, we reduce the number of features by defining groups of features that get altered together.
The past load is split into seven groups of 24 features, representing the past seven days.
All features belonging to the same exogenous time series form one feature group.
This makes 13 feature groups in total for the real-world data, and 14 feature groups in total for the synthetic data which exhibits one exogenous feature more.
%Consequently, it becomes feasible to use ten permutations instead of just two.
As a result, the model is called $13 \times 10 \times 2 \times 100 = 26\,000$ times to create a local explanation of a forecast on the real-world data.

The Custom Masker samples each feature group differently, respecting the characteristics of the feature and creating more realistic samples than the Permutation Explainer.
For a load feature group, 24 consecutive values from the training data are sampled, so that these 24 values form a smooth and realistic curve in themselves (but not necessarily realistic in the context of the other load values and exogenous features). 
The month values are increased by a constant, randomly chosen offset between 0 and 11 months.
In the same way, the day-of-week and hour-of-day values are offset by 0 to 6 days, respectively 0 to 23 hours. For the temperature feature, the entire temperature curve is replaced by 336 consecutive values from the training data.

\subsubsection*{From local to global explanations}

The local explanations of $m$ samples are aggregated to feature importance values by summing up the absolute SHAP values of a feature group in all samples, and then calculating percentages on the summed up values.

Another form of global model explanation are the dependence plots shown in Figures \ref{fig:global-explanations-synthetic} and \ref{fig:global-explanations-transnet}.
For these, $m$ samples result in $m$ dots, each with the feature value on the x-axis and the SHAP value on the y-axis. One explanation results in 168 SHAP values for the 168 predicted time steps, but only the SHAP values for the 24-hour forecast (i.e. the 24th value in the forecast horizon) are shown in the dependence plots.
Similarly, there are multiple feature values that could be shown on the x-axis (168 for the past load and 336 for each of the covariates), of which we chose the load of one week before the predicted time step for the load plot, and the future value of the predicted time step for all other plots.

\subsection{SHAPformer}
\label{sec:shapformer}

\subsubsection*{Attention manipulation}

As the main building block of the Transformer \cite{vaswani_attention_2017}, the attention mechanism \cite{bahdanau_neural_2015} computes a context-dependent embedding of a query vector given an arbitrary number of equally-dimensioned key vectors as context.
More formally, there is a query vector $q$ and key vectors $k_1$ to $k_n$.
With each key vector $k_i$ there is a value vector $u_i$ associated.
The attention score $a_i$ of a key vector $k_i$ is computed as $a_i = \dfrac{q^{T}k_i}{\sqrt{d}}$, where $d$ is the vector dimension.
The attention scores $a_i$ to $a_n$ are then soft-maxed by the formula $\alpha_i = \dfrac{\mathrm{exp}(a_i)}{\sum_{j=1}^n \mathrm{exp}(a_j)}$ in order to compute attention weights $\alpha_1$ to $\alpha_n$ which sum up to one.
The output $o$ of the attention mechanism is finally computed as $o = \sum_{i=1}^n \alpha_i \cdot u_i$.

Attention manipulation \cite{deiseroth_atman_2023} allows dropping out individual vectors $k_i$ by setting their attention score to $a_i = -\infty$, so that the attention weight $\alpha_i$ becomes zero after the softmax operation.
This means that the weight of vector $u_i$ in the weighted sum is set to zero, and therefore no information from $u_i$ is contained in the attention output and the following layers of the model have no access to $u_i$.

In the following, we make use of attention manipulation to restrict the model's access to feature groups that are absent in a coalition.

\subsubsection*{SHAPformer architecture}

\begin{figure}
    \hspace{-0.5cm}A SHAPformer architecture\\
    \makebox[\linewidth][c]{
    \begin{subfigure}{1.2\linewidth}
        \centering
        \includegraphics{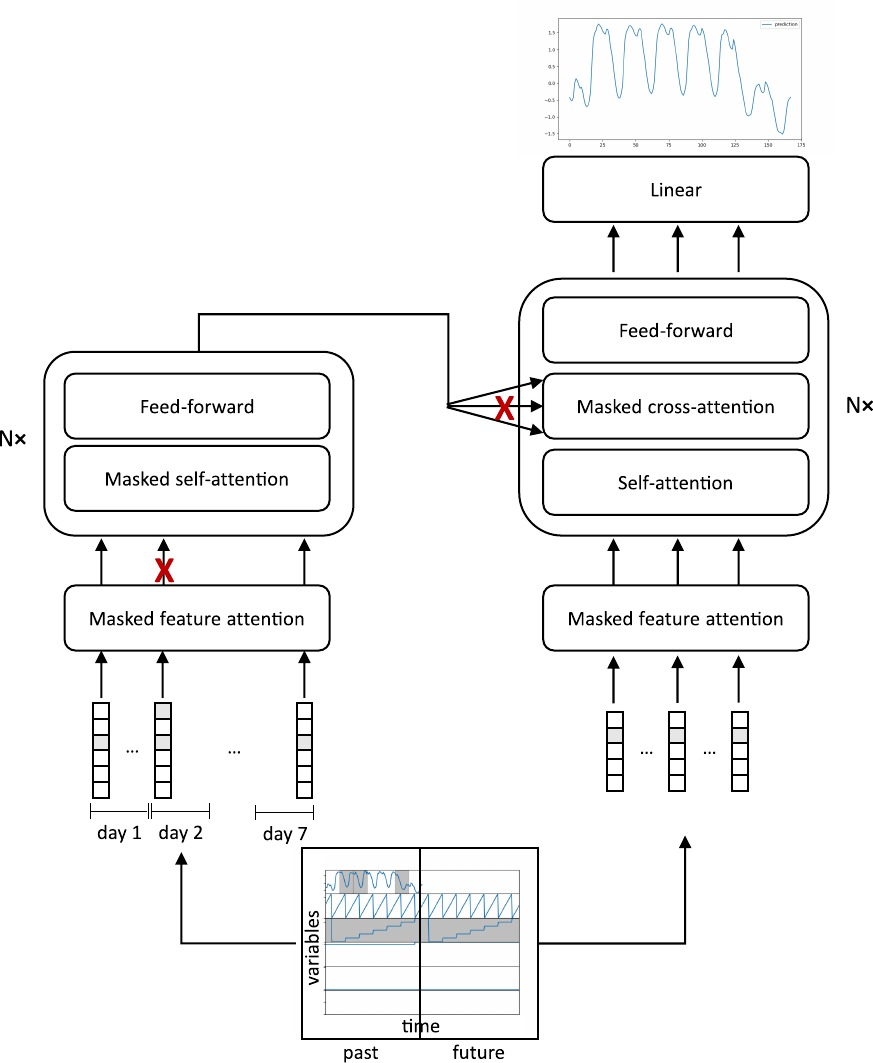}
    \end{subfigure}
    }

    \hspace{-0.5cm}B Masked feature attention\\
    \makebox[\linewidth][c]{
    \begin{subfigure}{1.2\linewidth}
        \centering
        \includegraphics[scale=0.5]{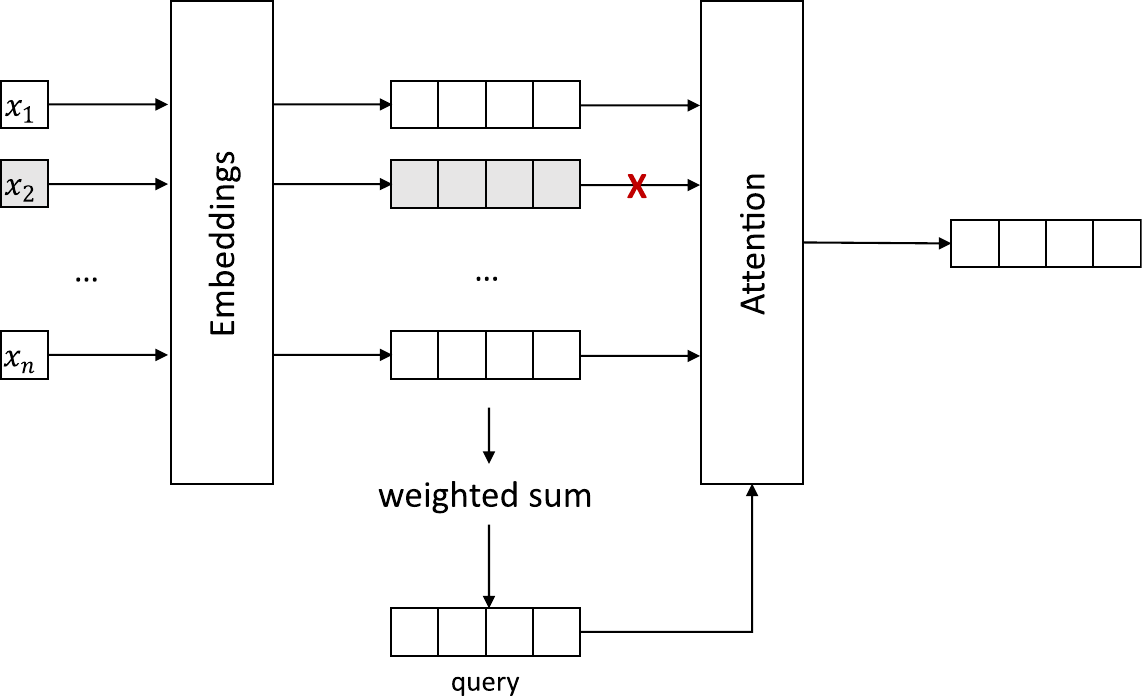}
    \end{subfigure}
    }
    \caption{Overview of the SHAPformer architecture (A) and the masked feature attention mechanism (B). $N$ encoder and decoder blocks are stacked. The masked feature attention is used to drop out variables ($v_2$ in the example in B). Masked self-attention and masked cross-attention are used to drop out time steps (days 2, 3 and 6 in the example in A).}
    \label{fig:shapformer-architecture}
\end{figure}

Figure \ref{fig:shapformer-architecture}A shows the SHAPformer architecture.
The model has a lookback size of 168 and a forecast horizon of 168.
The 168 past feature vectors are given to the encoder. Each past vector contains the load feature and exogenous features. The 168 future feature vectors are fed to the decoder. Each future vector contains the exogenous features.

SHAPformer uses a masked feature-attention mechanism to compute time step embeddings, as shown in Figure \ref{fig:shapformer-architecture}B.
This mechanism computes an embedding of the feature values belonging to one time step.
For each variable, the model maintains a learned embedding.
For categorical variables with $k$ possible values, $k$ randomly initialized embedding vectors are used.
For continuous variables, a linear layer is used to embed the feature values.
A query vector is computed as a weighted sum of the embedded variables, where dropped-out features get zero weight and all others weight one.
Then, attention is used to create the final embedding, based on the query vector and the embedded feature values as key vectors.
In the masked feature-attention mechanism, the attention weight of absent features is set to zero, so that the model has no access to their embeddings.
With this approach, SHAPformer can be run on subsets of the exogenous features.
A positional encoding is added to the embedding vectors to maintain their order.

For the load feature, masked self-attention in the encoder and masked cross-attention in the decoder are used to restrict access to 24 time steps belonging to the same load feature group.

The encoder gets 168 embeddings for the 168 past time steps as input.
Multiple Transformer encoder layers are stacked, each consisting of self-attention and linear layers -- both equipped with add and norm mechanisms (not shown in the figure).
The self-attention is masked, i.e. the attention value for absent time steps is set to zero, so that they are not used to calculate the attention output.

The decoder gets 168 vectors for the 168 future time steps as input.
It consists of multiple Transformer decoder layers using self-attention and cross-attention to access the encoder output.
The cross-attention is masked, i.e. the attention value for absent time steps is set to zero, so that the decoder has no access to them.
Finally, a linear layer is used as prediction head to transform the 168 output vectors of the decoder into 168 scalar values, which are the model prediction for the next 168 time steps.

During model training, random masks are sampled for every example that is given to the model, so that each exogenous feature group and each day has a 50\% chance of being masked.
Thereby, the model learns to create robust forecasts using subsets of the features. Without masked training, the model could potentially behave arbitrarily when a feature is absent at inference time.

\subsubsection*{Owen values}

Owen values \cite{owen_values_1977} are a variant of SHAP values for games with a coalitional structure.
First, the contributions of the coalitions are computed, and then they are broken down further into contributions of the individual players.
The formula is slightly different from the formula for SHAP values:
\begin{equation}
\textit{Owen}(v_i) = \sum_{R \subseteq M \setminus \{k\}} \sum_{T \subseteq B_k \setminus \{v_i\}} \dfrac{1}{|M| \cdot |B_k|} \dfrac{1}{\binom{|M|-1}{|R|}} \dfrac{1}{\binom{|B_k|-1}{|T|}}(f(Q \cup T \cup \{v_i\}) - f(Q \cup T))
\end{equation}
with $M=\{B_1, ... B_l\}$ being the set of coalitions, $B_k$ the coalition containing $v_i$, and $Q=\bigcup_{r \in R} B_r$ the union of features over a subset of coalitions $R$.
We use the Owen value formula in SHAPformer in order to treat the past load equally as the exogenous features on the coalition level, and then break the past loads' SHAP value down further into SHAP values of the seven input days.

Note that in our case, when the load of a past day is absent, the other features of that day are also absent, as an effect of the masked self-attention in the encoder and masked cross-attention in the decoder.
We assume that this does not affect the Owen values, because the past day's features are used as context information for the past day's load -- when the load is not available to the model, the effect of the past day's features on the forecast is negligible.

\subsection{Synthetic dataset}
\label{sec:synthetic-dataset}

\subsubsection*{Data generation}

In order to verify the explanations from SHAPformer, we create a large synthetic dataset for time-series forecasting.
Since the data generation process is known, it is possible to judge the explanations returned from SHAPformer and other XAI methods.

The dataset consists of 120\,000 examples. Each example contains two weeks of hourly values from the target variable and multiple covariates.
The first week of the target variable is used as input to the model, and the model has to predict the second week of the target variable.
We call the target variable "load", as in our real dataset.
For the covariates, both weeks are used as input to the model.
That is, we assume a perfect forecast of the covariates for the next week.

The generation of an example follows a multi-step procedure:
\begin{itemize}
    \item A month $\in [1, 12]$, start weekday $\in [0, 6]$ and start hour $\in [0, 23]$ are sampled with uniform probability.
    \item A base load is sampled uniformly in the range $(-0.5, 0.5)$.
    \item The month has an additive effect on the base load. A value of $0.1 \cdot \mathrm{sin}(\mathrm{month} \cdot 2 \pi / 12)$ gets added to the base load.
    \item A daily load curve of 24 values is generated as $\mathit{factor} \cdot (\mathrm{sin}(h - 0.5 \cdot \pi) + \alpha \cdot \mathrm{sin}(h \cdot 2\pi / 24) + \beta \cdot \mathrm{cos}(h \cdot 2\pi / 24))$, with $\mathit{factor} \in (0.5, 1)$, $\alpha, \beta \in (-0.5, 0.5)$ uniformly sampled.
    \item A uniformly sampled pattern of 24 values with mean 0 and standard deviation 0.1 is added to the daily load. We call the result the $\mathrm{day\_pattern}$ and use it as the load curve for workdays.
    \item The Saturday pattern deviates from the workday pattern by a multiplicative factor and an additive deviation. It is generated as $s_1 \cdot \mathrm{day\_pattern} + \mathrm{normal}(0, 0.1)$, with $s_1 \in (0.5, 0.9)$ uniformly sampled, and $\mathrm{normal(0, 0.1)}$ generating 24 values from a normal distribution with mean 0 and standard deviation 0.1.
    \item The Sunday pattern is $s_2 \cdot \mathrm{day\_pattern}  + \mathrm{normal}(0, 0.1)$, with $s_2 \in (0.2, s_1)$.
    \item From the workday, Saturday and Sunday patterns, the 336 hourly values are produced by repeating the daily pattern, Saturday pattern and Sunday pattern, beginning from the start day of week and start hour of day. Each day has a 10\% chance of being a holiday. On holidays, the Sunday pattern is always used, independent of the weekday.
    \item A temperature curve is generated by a random walk starting from a uniformly sampled value in (-0.5, 0.5) and each step sampled from a normal distribution with mean 0 and standard deviation 0.02. The temperature has a multiplicative effect on the load. That is, the load is rescaled by the temperature as $\mathrm{load} \cdot (0.5 + 0.5 \cdot \mathrm{temperature})$.
    \item Finally, random noise is added to the load time series, sampled from a normal distribution with mean 0 and standard deviation 0.05.
\end{itemize}

The following covariates are used as inputs for the forecasting models: hour of day, day of week, month (all categorical), holiday (binary), temperature and two uncorrelated noise features (all continuous).

100\,000 examples are used for training, 10\,000 for validation and 10\,000 for test.
Two examples are shown in Figure \ref{fig:local-explanations-synthetic}A.

\subsubsection*{Ground truth explanations}

In order to validate the explanations from the different XAI methods presented in Section \ref{sec:results-synthetic}, we calculate ground truth SHAP values for the synthetic dataset.
This is possible because we have access to the data generation procedure and can therefore explain the true data dependencies with SHAP.
As SHAP is model agnostic, it can be used to explain an arbitrary function $f(x)$. Usually, in the context of XAI, $f$ is a machine learning model, but we set $f$ to the data generation process in order to calculate ground truth explanations.
The inputs $x$ to the data generation process consist of the past load, hour of day, day of week, month, holiday, multiplier and two noise features, each represented as a time series of 168 past and -- for all inputs except for the load -- 168 future values, and the output is the target load of the next week.

We compute ground truth explanations for the examples in the test set of the synthetic dataset.
To do so, we use the SHAP Permutation Explainer on the data generation process.
For a given test example and a permutation, the Permutation Explainer computes marginal contributions of the inputs based on subsets of the inputs.
The inputs that are absent, i.e. not contained in the subset, are resampled 1000 times following the sampling process described above, and an alternative target load curve is generated using the data generation process and the resampled inputs.
These 1000 alternative targets are then averaged, and the averaged values are used as the expected target load curve given the subset of the inputs.
From the target load curves generated with different input subsets, marginal contributions and SHAP values are computed with the procedure described in Section \ref{sec:shap}.

While sampling alternative inputs, we make sure that no unrealistic combinations of inputs are generated.
In particular, when an input $x_1$ depends on an input $x_2$ that is contained in the active subset, we do not resample $x_1$.
The following dependencies are respected:
If the day of the week is in the active subset, the hour of day is not resampled (as it can be inferred from the day beginnings and endings).
If the holiday feature is in the active subset and there is at least one holiday in the example, the hour of day is not resampled (for the same reason).
If the load of the past week is in the active subset, all calendric information as well as the multiplier of the last week is not resampled (because they affect the load), as well as the load patterns for the workday, Saturday and Sunday (as changing any of these would affect the past load).

Note that the ground truth SHAP values are meant to be true to the data \cite{chen_true_2020} under the assumptions stated above. However, a machine learning model could learn different patterns,
%(e.g. not inferring the hour of day feature from the day of week feature)
so that explanations being true to the model instead of true to the data would deviate from our ground truth.

\subsection{Baselines}
\label{sec:baselines}

\subsubsection*{Persistence baseline}
The persistence baseline predicts the value from one week before the predicted time step. It thereby respects daily and weekly seasonalities, but no exogenous dependencies. In its simplicity, it is inherently interpretable.

\subsubsection*{Linear Regression}
The linear regression model is based on the following features: the load from 168 time steps before the predicted time step, the hour of day (sine and cosine encoded), the day of week (sine and cosine encoded), the month (sine and cosine encoded), whether it is a holiday (binary) and the temperature and precipitation values.
As the model is linear, an explanation can be derived from the model coefficients.

\subsubsection*{XGBoost Regressor}
The Extreme Gradient Boosting Regressor (XGBoost) uses the same features as the linear regression. The default hyperparameters, 100 estimators and a maximum depth of three, are used.
The model is able to return feature importance values, but no information on how the features impact the prediction is available.

\subsubsection*{Time-series Transformer}
The time-series Transformer is the same model architecture as SHAPformer, but the model is trained without masking, so that the model always receives the full information about the last week and exogenous features for the next week.
The Permutation Explainer and Custom Masker are used with the time-series Transformer to generate explanations.

% We use an encoder-decoder Transformer for time-series forecasting as in \cite{hertel_transformer_2023}.
% The context length and the forecast horizon are both 168 hours.
% The encoder gets 168 vectors for the 168 past time steps as input.
% Each vector contains the load value and $n$ exogenous variables.
% The decoder gets 168 vectors for the 168 future time steps as input, each comprising the $n$ exogenous variables.
% The vectors are embedded with feature-wise attention (see Section \ref{sec:shapformer-architecture}).

\subsubsection*{Temporal Fusion Transformer}
The Temporal Fusion Transformer \cite{lim_temporal_2021} is an encoder-decoder model based on an LSTM encoder and an LSTM decoder \cite{hochreiter_long_1997}, followed by a single Transformer \cite{vaswani_attention_2017} layer. It uses causal masking in the Transformer layer to prevent the model from using information of time steps after the prediction time step.
We use the same features as for SHAPformer for both the encoder and the decoder, and no static covariates.
Crucially, the model's first layer is a variable selection network, which computes two sets of variable weights -- one for the encoder and one for the decoder -- that can be interpreted as feature importance values.

% \subsubsection{Experiment repetitions}

% We noted some variance in the results on the real-world data when repeating the experiments. The numbers in the results table for the Transformer, TFT and SHAPformer on the real-world data are averages from five repetitions. The figures are created with the best model from the five repetitions.

\backmatter

%\section*{Supplementary information}
\bmhead{Supplementary information}

For supplementary information on the data and methods, as well as detailed results from the comparison methods, please refer to the Appendix.

% If your article has accompanying supplementary file/s please state so here. 

% Authors reporting data from electrophoretic gels and blots should supply the full unprocessed scans for key as part of their Supplementary information. This may be requested by the editorial team/s if it is missing.

% Please refer to Journal-level guidance for any specific requirements.

\bmhead{Acknowledgements}

During the preparation of this work, the authors used ChatGPT 4o to improve the clarity and fluency of the written text. After using this tool, the authors reviewed and edited the content as needed and take full responsibility for the content of the published article.

% Acknowledgements are not compulsory. Where included they should be brief. Grant or contribution numbers may be acknowledged.

% Please refer to Journal-level guidance for any specific requirements.

\section*{Declarations}

% Some journals require declarations to be submitted in a standardised format. Please check the Instructions for Authors of the journal to which you are submitting to see if you need to complete this section. If yes, your manuscript must contain the following sections under the heading `Declarations':

% \begin{itemize}
% \item Conflict of interest/Competing interests (check journal-specific guidelines for which heading to use)
% \item Ethics approval and consent to participate
% \item Consent for publication
% \item Materials availability
% \end{itemize}

% \noindent
% If any of the sections are not relevant to your manuscript, please include the heading and write `Not applicable' for that section. 

\subsection*{Funding}

The authors gratefully acknowledge funding by the Helmholtz Association under the program “Energy System Design” and the Helmholtz Association’s Initiative and Networking Fund through Helmholtz AI.

\subsection*{Data availability}

The real-world load data is publicly available and downloadable via \href{https://open-power-system-data.org}{open-power-system-data.org} \cite{wiese_open_2019}.
The exogenous weather data is publicly available and downloadable via \href{https://cds.climate.copernicus.eu/datasets/sis-energy-derived-reanalysis}{cds.climate.copernicus.eu} \cite{copernicus_climate_change_service_climate_2020}.
The synthetic dataset with ground truth explanations is available via GitHub under \url{https://github.com/KIT-IAI/SHAPformer}.

\subsection*{Code availability}

SHAPformer is available as a Python package via GitHub under \url{https://github.com/KIT-IAI/SHAPformer}.

\subsection*{Author contribution}

\textbf{M.H.:} Conceptualization, Methodology, Software, Investigation, Writing - original draft, Visualization;
\textbf{S.P.:} Methodology, Writing - review \& editing;
\textbf{R.M.:} Conceptualization, Supervision, Funding acquisition, Writing - review \& editing;
\textbf{V.H.:} Supervision, Funding acquisition, Writing - review \& editing;
\textbf{B.S.:} Conceptualization, Funding acquisition, Writing - review \& editing.

% \bigskip

% Springer journals and proceedings: \url{https://www.springer.com/gp/editorial-policies}

% Nature Portfolio journals: \url{https://www.nature.com/nature-research/editorial-policies}

% \textit{Scientific Reports}: \url{https://www.nature.com/srep/journal-policies/editorial-policies}

% BMC journals: \url{https://www.biomedcentral.com/getpublished/editorial-policies}
% \end{flushleft}

\bibliography{zotero-references}% common bib file

% \clearpage
\begin{appendices}

\section{Detailed results on the real-world dataset}
\label{app:detailed-results-transnet}

Detailed results on the empirical dataset from TransnetBW are given in Table \ref{tab:detailed-results-transnet}.
Five metrics are evaluated and means and standard deviations from five runs are reported.
The baseline, linear regression and XGBoost are deterministic, therefore they have a standard deviation of zero.

\begin{table}[hbt]
    \caption{Forecast errors on the TransnetBW dataset, evaluated with five different metrics. Means and standard deviations over five runs are reported.}
    \label{tab:detailed-results-transnet}
    \centering
    \begin{tabular}{lrrrrr}
        \toprule
        Model & \multicolumn{1}{c}{MAE} & \multicolumn{1}{c}{MSE} & \multicolumn{1}{c}{MAE} & \multicolumn{1}{c}{RMSE} & \multicolumn{1}{c}{MAPE} \\
          & \multicolumn{1}{c}{(scaled)} & \multicolumn{1}{c}{(scaled)} & \multicolumn{1}{c}{[MW]} & \multicolumn{1}{c}{[MW]} & \multicolumn{1}{c}{[\%]} \\
        \midrule
        Persistence baseline & $0.255 \pm 0.000$ & $0.177 \pm 0.000$ & $395.5 \pm \phantom{0}0.0$ & $652.3 \pm \phantom{0}0.0$ & $6.17 \pm 0.00$ \\
        Linear Regression & $0.255 \pm 0.000$ & $0.128 \pm 0.000$ & $395.0 \pm \phantom{0}0.0$ & $553.7 \pm \phantom{0}0.0$ & $6.19 \pm 0.00$ \\
        XGBoost & $0.177 \pm 0.000$ & $0.062 \pm 0.000$ & $274.2 \pm \phantom{0}0.0$ & $387.0 \pm \phantom{0}0.0$ & $4.18 \pm 0.00$ \\
        TFT & $0.163 \pm 0.016$ & $0.065 \pm 0.016$ & $251.9 \pm 24.3$ & $390.8 \pm 49.9$ & $3.94 \pm 0.41$ \\
        Transformer & $0.127 \pm 0.005$ & $0.028 \pm 0.002$ & $197.3 \pm \phantom{0}8.3$ & $263.1 \pm \phantom{0}9.1$ & $2.98 \pm 0.12$ \\
        SHAPformer & $0.131 \pm 0.006$ & $0.029 \pm 0.002$ & $203.3 \pm \phantom{0}8.9$ & $265.9 \pm \phantom{0}9.6$ & $3.09 \pm 0.13$ \\
        \bottomrule
    \end{tabular}
\end{table}

\section{Hyperparameter optimization}
\label{sec:hyperparameters}

The hyperparameters of the Temporal Fusion Transformer, the Transformer and SHAPformer on synthetic data were optimized using Bayesian Optimization and Weights and Biases (wandb) \cite{biewald_experiment_2020}. 
For SHAPformer on real-world data, the same hyperparameters were used as for the Transformer, but with a lower learning rate, which stabilizes the masked training.
%For XGBoost, the default hyperparameters were chosen.
The selected hyperparameters are given in Table \ref{tab:hyperparameters}.
Adam \cite{kingma_adam_2017} and AdamW \cite{loshchilov_decoupled_2019} were used as training algorithms, using the mean squared error loss function.

\begin{table}[hbt]
    \caption{The chosen hyperparameters for the different models.}
    \label{tab:hyperparameters}
    \begin{tabular}{lccc}
        \toprule
        Model & Hyperparameter & Synthetic & Real \\
        \midrule
        Temporal Fusion Transformer & $d\_model$ & 256 & 1024 \\
                & batch size & 4 & 16 \\
                & heads & 8 & 8 \\
                & optimizer & AdamW & AdamW \\
                & learning rate & 0.00067 & 0.00017 \\
                & decay rate & 0.37 & 0.93 \\
        \hdashline
        Transformer & layers & 8 & 2 \\
                & $d\_model$ & 64 & 128 \\
                & heads & 4 & 2 \\
                & optimizer & AdamW & Adam \\
                & batch size & 16 & 64 \\
                & learning rate & 0.00015 & 0.00010 \\
                & decay rate & 1.00 & 1.00 \\
        \hdashline
        SHAPformer & layers & 7 & 2 \\
                & $d\_model$ & 512 & 128 \\
                & heads & 2 & 2 \\
                & optimizer & AdamW & Adam \\
                & batch size & 16 & 64 \\
                & learning rate & 0.00010 & 0.00001 \\
                & decay rate & 0.96 & 1.00 \\
        \bottomrule
    \end{tabular}
\end{table}

\section{Local SHAPformer explanations on synthetic data}
\label{app:local-explanations}

\begin{figure}
    \hspace{-1.5cm}A Synthetic examples\\
    \makebox[\linewidth][c]{
    \begin{subfigure}{0.8\linewidth}
        \centering Feature groups \& target \\
        \includegraphics[width=\textwidth]{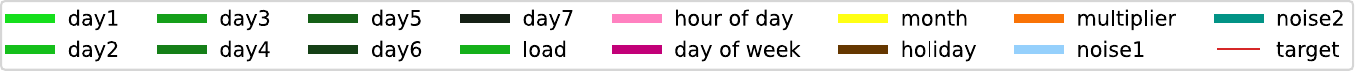}
    \end{subfigure}
    }
    \vspace{-1cm}
    
    \makebox[\linewidth][c]{
    \begin{subfigure}{0.6\linewidth}
        \includegraphics[width=\textwidth]{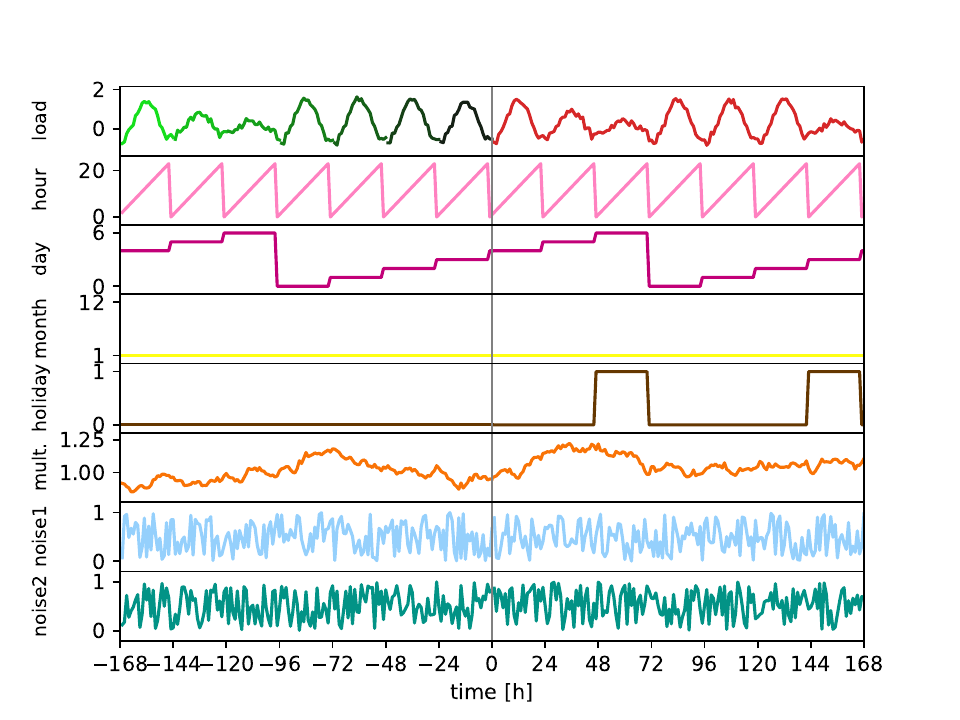}
    \end{subfigure}
    \begin{subfigure}{0.6\linewidth}
        \includegraphics[width=\textwidth]{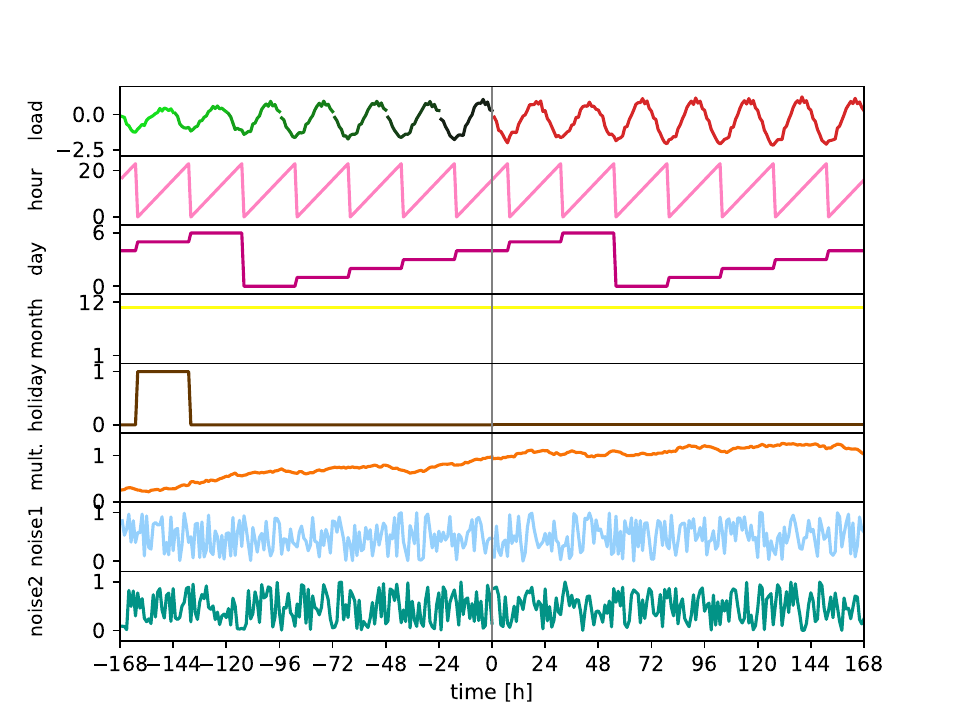}
    \end{subfigure}
    }

    \hspace{-1.5cm}B SHAPformer explanations\\
    \makebox[\linewidth][c]{
    \begin{subfigure}{0.6\linewidth}
        \includegraphics[width=\textwidth]{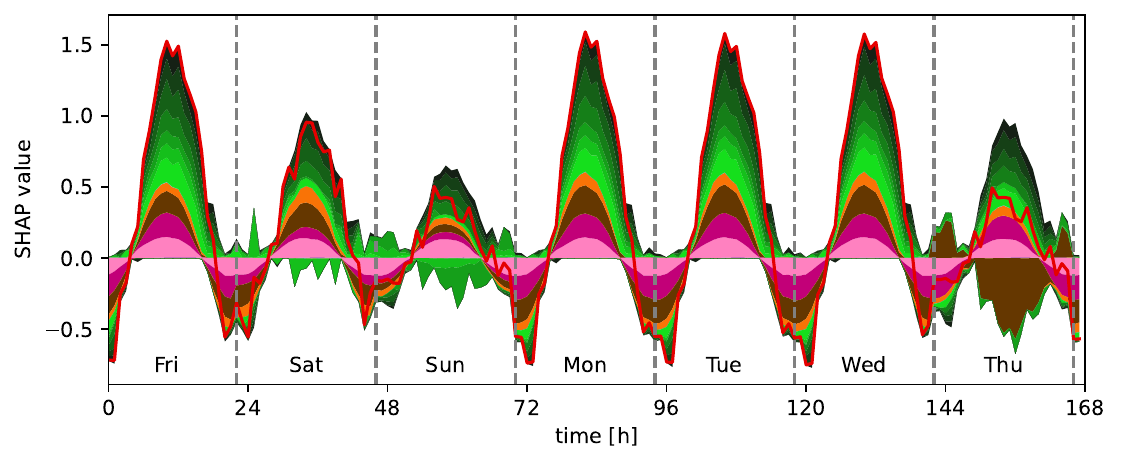}
    \end{subfigure}
    \begin{subfigure}{0.6\linewidth}
        \includegraphics[width=\textwidth]{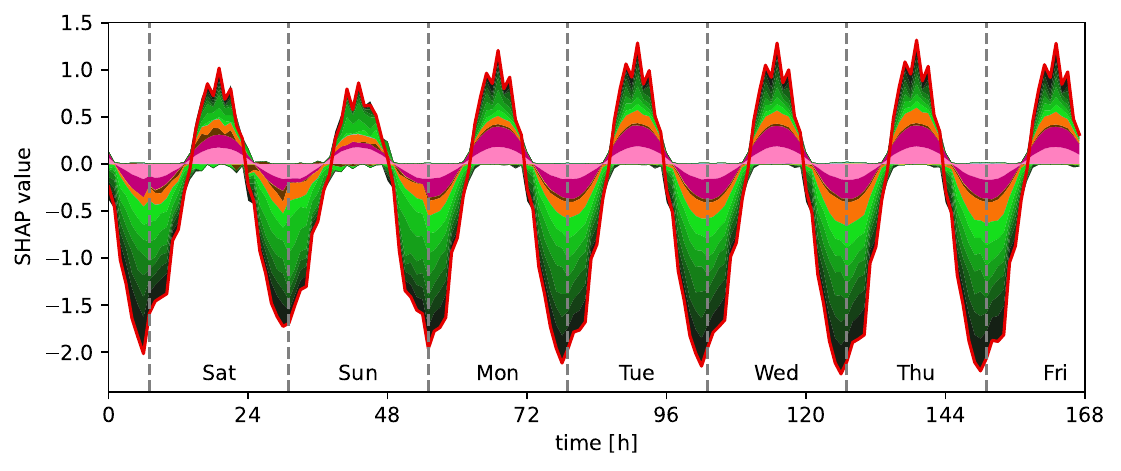}
    \end{subfigure}
    }
    
    \hspace{-1.5cm}C Ground truth\\
    \makebox[\linewidth][c]{
    \begin{subfigure}{0.6\linewidth}
        \includegraphics[width=\textwidth]{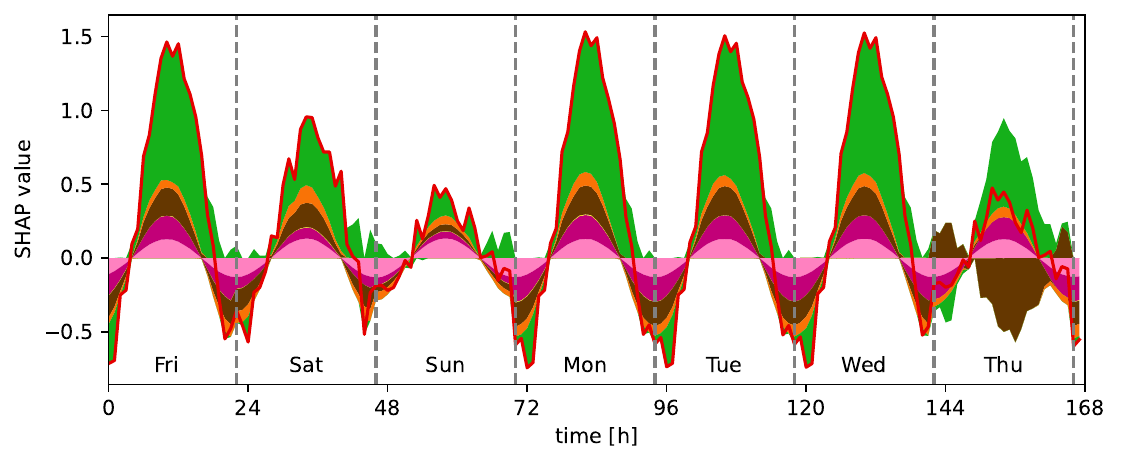}
    \end{subfigure}
    \begin{subfigure}{0.6\linewidth}
        \includegraphics[width=\textwidth]{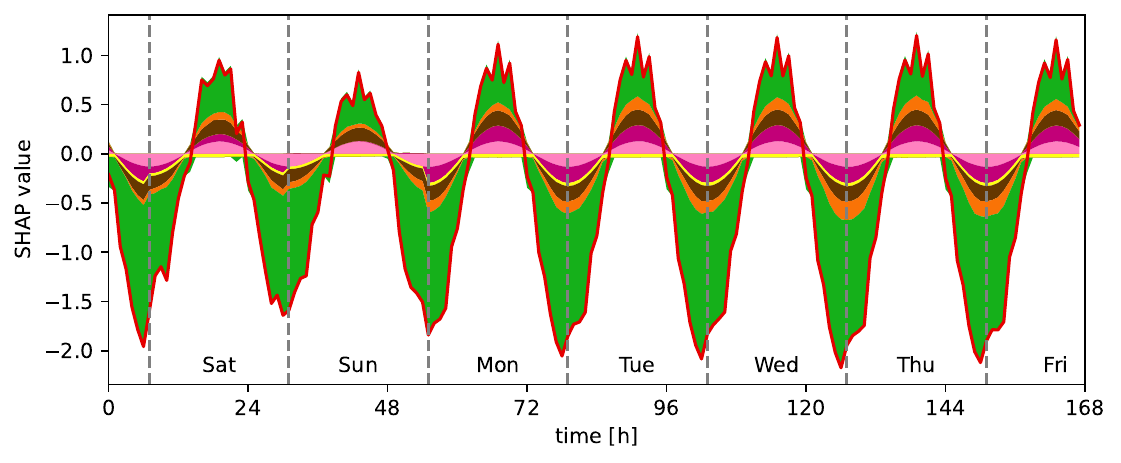}
    \end{subfigure}
    }

    \caption{Local explanations of two different synthetic examples (left and right).
    Left: the holiday feature affects the last predicted day. Right: The increasing multiplier has an increasing effect.}
    \label{fig:local-explanations-synthetic}
\end{figure}

Local explanations of two synthetic examples are shown in Figure \ref{fig:local-explanations-synthetic}.
On the left-hand side, the forecast horizon starts on a Friday.
%The hour of day, day of week and holiday features influence the prediction in a way that resembles the base load curve: half of a sine curve with higher loads during the day than at night. This Sunday is a holiday and the holiday feature and day of week features have only a small effect compared to the weekdays -– i.e., the load on this Sunday is lower. The previous Sunday (input day 3) has a negative effect on the predicted Sunday -– i.e., because the previous Sunday had a particularly low load, the prediction on the next Sunday is lowered. The last day of the forecast horizon is a holiday and the holiday feature has a shrinking effect, i.e. decreasing the positive loads during the day and increasing the negative loads at night.
The hour of day, day of week, and holiday features influence the prediction in a pattern resembling the base load curve: a half-sine shape with higher loads during the day and lower loads at night. In this case, Sunday is also a holiday, so the holiday and day-of-week effects are both small compared to typical weekdays, resulting in a lower overall load. The previous Sunday (input day 3) exerts a negative influence on the predicted Sunday, as its particularly low load leads to a reduced forecast for the following week. The final day of the forecast horizon is also a holiday, where the holiday feature compresses the load curve—reducing positive daytime loads and amplifying negative nighttime loads.
The ground truth looks similar to the SHAPformer explanation. Only the effect of the seven days is summarized, so the effect of input day 3 on the Sunday is not visible.
On the right-hand side, the prediction starts on a Friday evening and the load is affected by a temperature increase. This is visible in the explanation of the forecast, where the temperature SHAP values increase in magnitude over time. In comparison to the ground truth, SHAPformer underestimates the month effect, as observable through the lower amplitudes in the explanation than in the ground truth for these two features.

\section{Transformer explained with SHAP}\label{app:shap-explanations}

This appendix shows the global and local explanations for the Transformer model created with the Permutation Explainer and the Custom Masker on the two datasets.

\subsection{Synthetic data}

\begin{figure}
    \hspace{-1.5cm}A Permutation Explainer\\
    \makebox[\linewidth][c]{
    \begin{subfigure}{0.4\linewidth}
        \includegraphics[width=\textwidth]{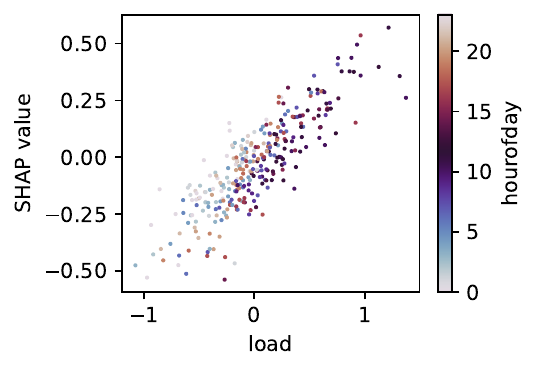}
    \end{subfigure}
    \begin{subfigure}{0.4\linewidth}
        \includegraphics[width=\textwidth]{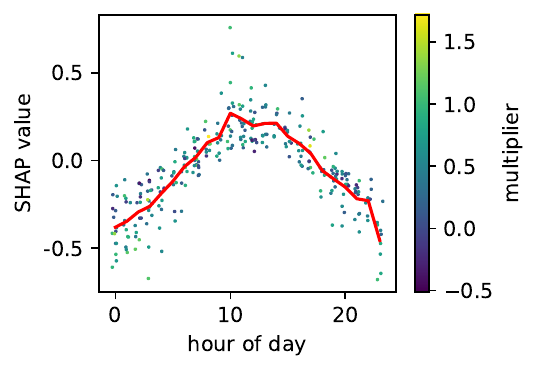}
    \end{subfigure}
    \begin{subfigure}{0.4\linewidth}
        \includegraphics[width=\textwidth]{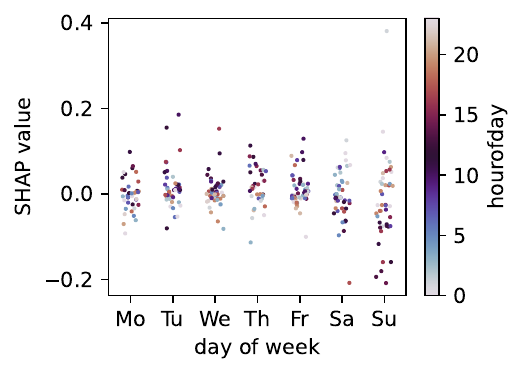}
    \end{subfigure}
    }
    \makebox[\linewidth][c]{
    \begin{subfigure}{0.4\linewidth}
        \includegraphics[width=\textwidth]{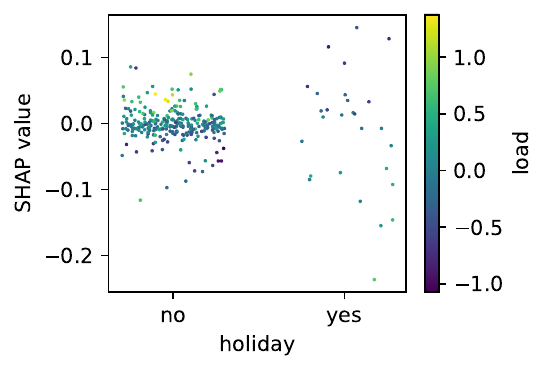}
    \end{subfigure}
    \begin{subfigure}{0.4\linewidth}
        \includegraphics[width=\textwidth]{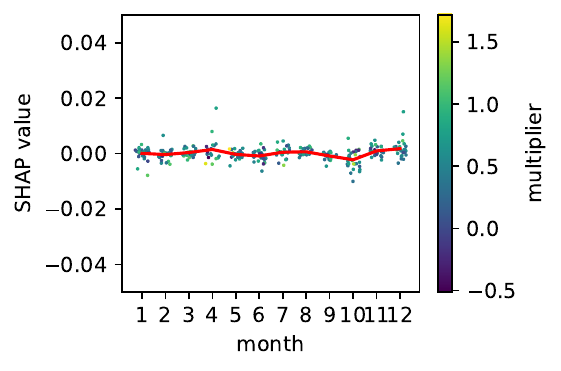}
    \end{subfigure}
    \begin{subfigure}{0.4\linewidth}
        \includegraphics[width=\textwidth]{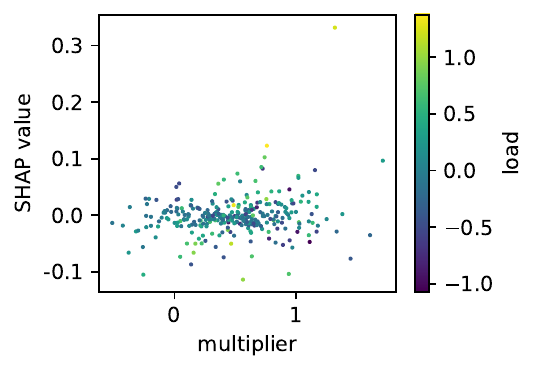}
    \end{subfigure}
    }
    
    \hspace{-1.5cm}B Custom Masker\\
    \makebox[\linewidth][c]{
    \begin{subfigure}{0.4\linewidth}
        \includegraphics[width=\textwidth]{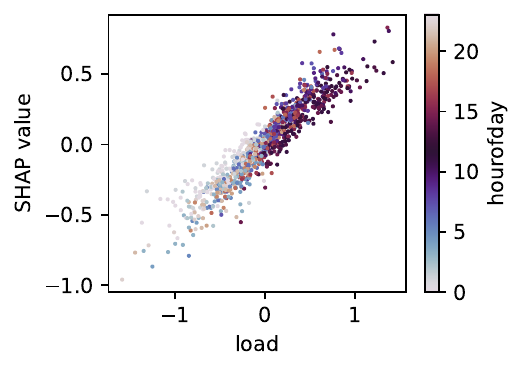}
    \end{subfigure}
    \begin{subfigure}{0.4\linewidth}
        \includegraphics[width=\textwidth]{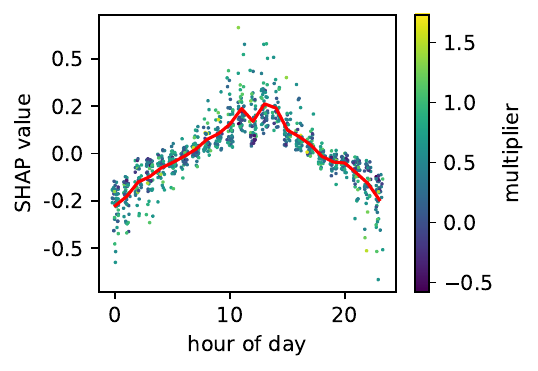}
    \end{subfigure}
    \begin{subfigure}{0.4\linewidth}
        \includegraphics[width=\textwidth]{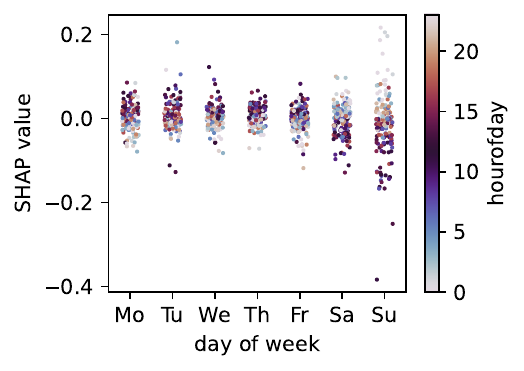}
    \end{subfigure}
    }
    \makebox[\linewidth][c]{
    \begin{subfigure}{0.4\linewidth}
        \includegraphics[width=\textwidth]{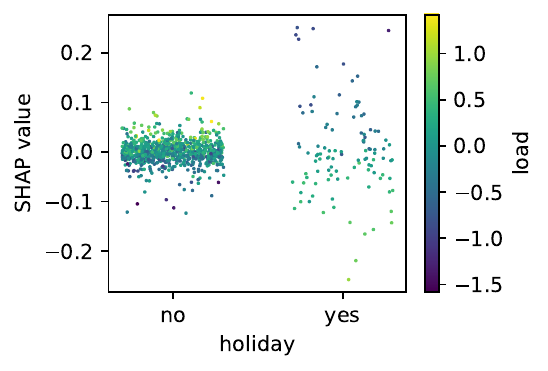}
    \end{subfigure}
    \begin{subfigure}{0.4\linewidth}
        \includegraphics[width=\textwidth]{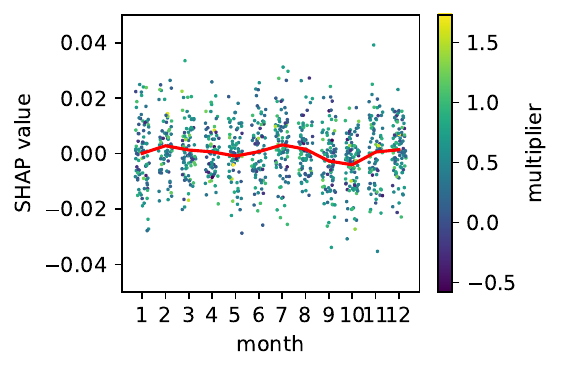}
    \end{subfigure}
    \begin{subfigure}{0.4\linewidth}
        \includegraphics[width=\textwidth]{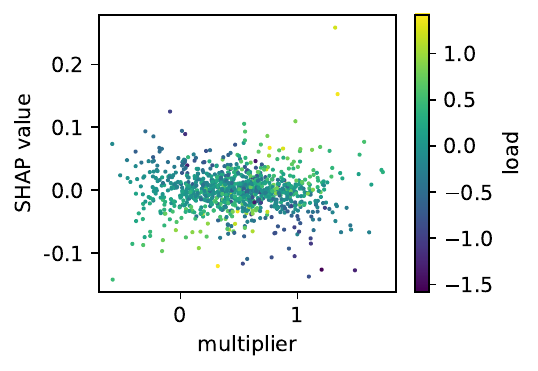}
    \end{subfigure}
    }
    
    \caption{Dependence plots on synthetic data created with the Permutation Explainer and the Custom Masker. For discrete variables, noise was added in the x-direction for visibility reasons.}
    \label{fig:global-explanations-synthetic-shap}
\end{figure}

\begin{figure}
    \hspace{-1.5cm}A Synthetic examples\\
    \makebox[\linewidth][c]{
    \hspace{1.5cm}
    \begin{subfigure}{0.8\linewidth}
        \includegraphics[width=\textwidth]{figures/legend_synthetic.pdf}
    \end{subfigure}
    }
    \vspace{-1cm}
    
    \makebox[\linewidth][c]{
    \begin{subfigure}{0.6\linewidth}
        \includegraphics[width=\textwidth]{figures/data/synthetic/test_35.pdf}
    \end{subfigure}
    \begin{subfigure}{0.6\linewidth}
        \includegraphics[width=\textwidth]{figures/data/synthetic/test_2753.pdf}
    \end{subfigure}
    }

    \hspace{-1.5cm}B Explanations from Permutation Explainer\\
    \makebox[\linewidth][c]{
    \begin{subfigure}{0.6\linewidth}
        \includegraphics[width=\textwidth]{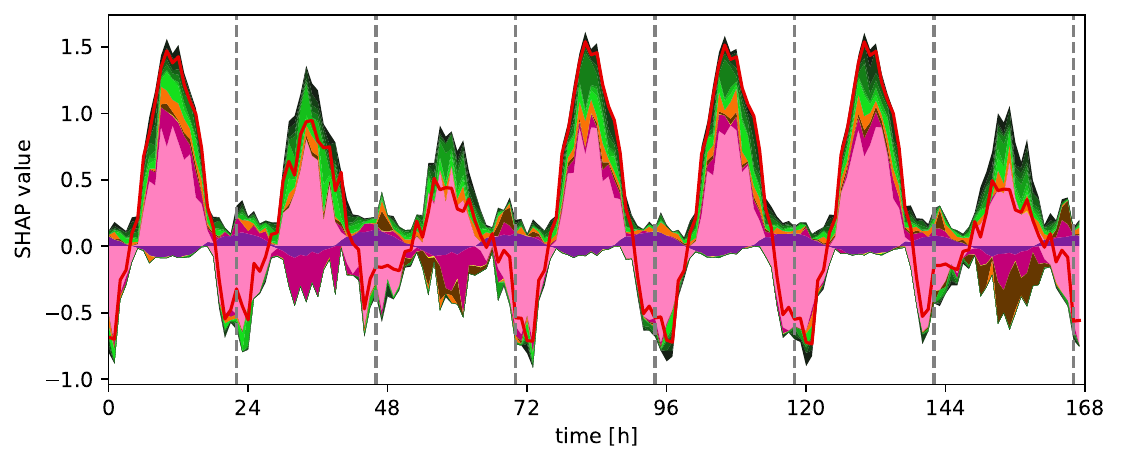}
    \end{subfigure}
    \begin{subfigure}{0.6\linewidth}
        \includegraphics[width=\textwidth]{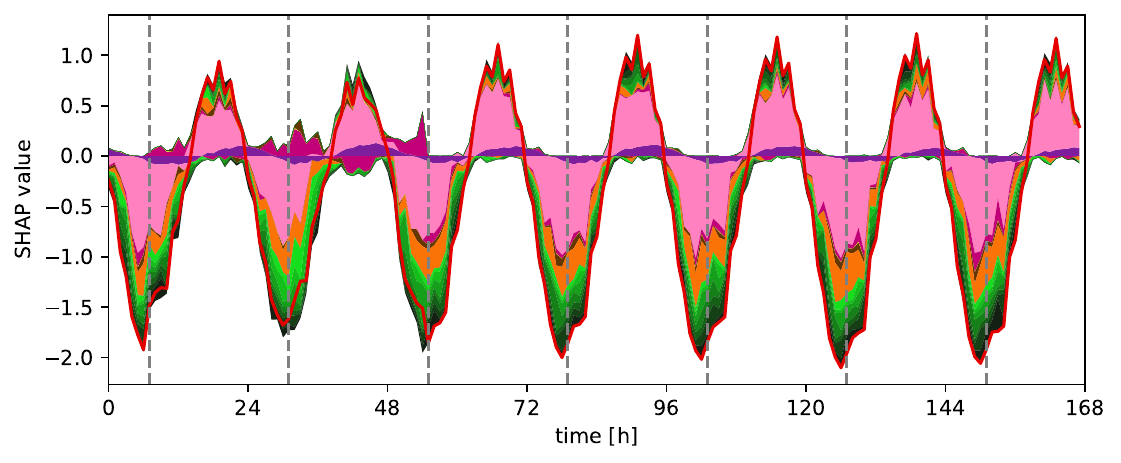}
    \end{subfigure}
    }
    
    \hspace{-1.5cm}C Explanations from Custom Masker\\
    \makebox[\linewidth][c]{
    \begin{subfigure}{0.6\linewidth}
        \includegraphics[width=\textwidth]{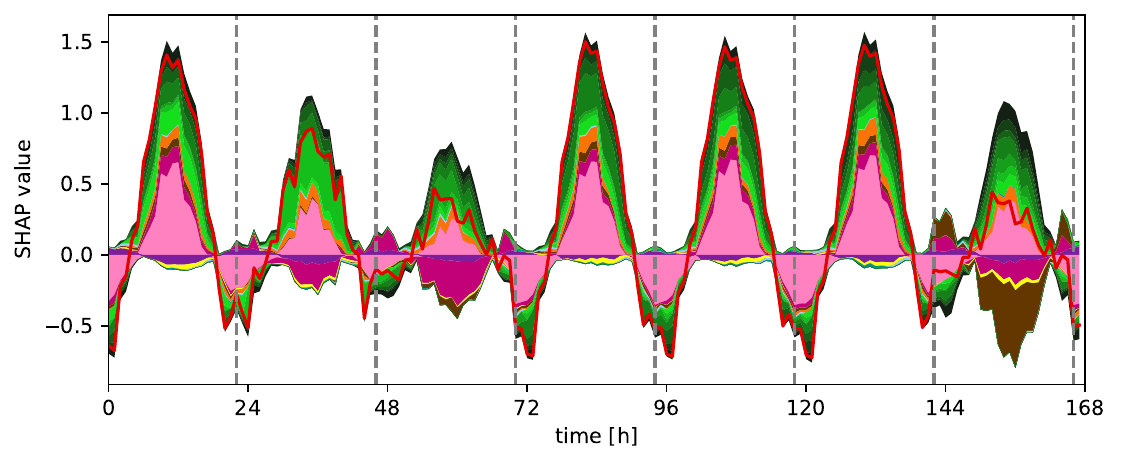}
    \end{subfigure}
    \begin{subfigure}{0.6\linewidth}
        \includegraphics[width=\textwidth]{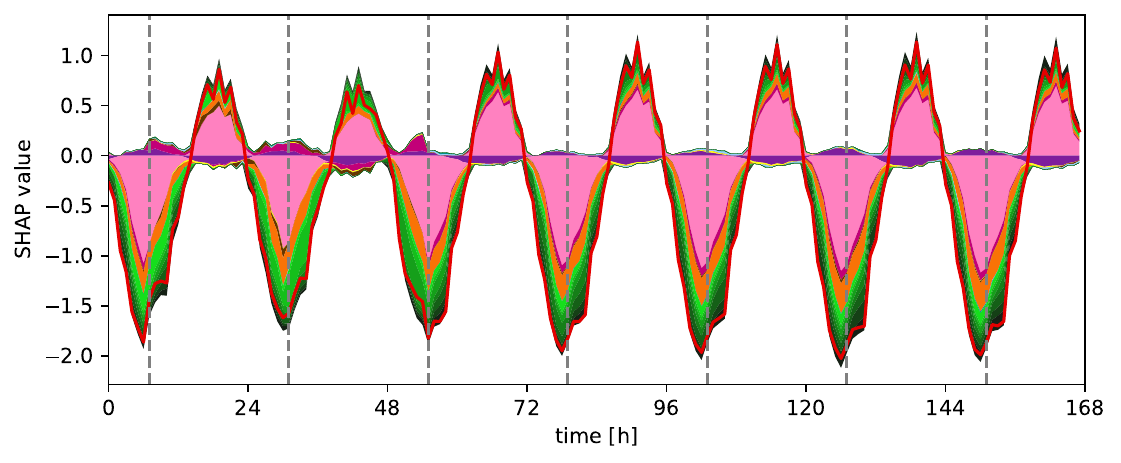}
    \end{subfigure}
    }
    
    \caption{Local explanations of synthetic examples created with the Permutation Explainer and the Custom Masker.}
    \label{fig:local-explanations-synthetic-shap}
\end{figure}

See Figure \ref{fig:global-explanations-synthetic-shap} for the global explanations and Figure \ref{fig:local-explanations-synthetic-shap} for the local explanations on the synthetic data.

\subsection{TransnetBW data}

\begin{figure}
    \hspace{-1.5cm}A Permutation Explainer\\
    \makebox[\linewidth][c]{
    \begin{subfigure}{0.4\linewidth}
        \includegraphics[width=\textwidth]{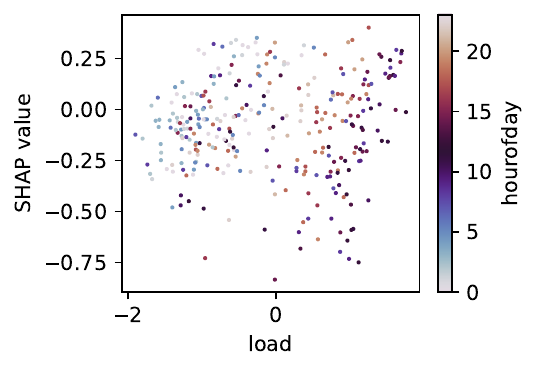}
    \end{subfigure}
    \begin{subfigure}{0.4\linewidth}
        \includegraphics[width=\textwidth]{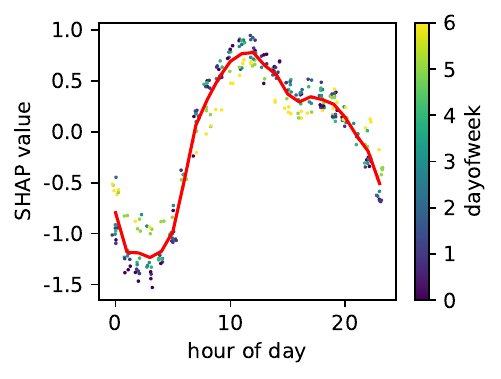}
    \end{subfigure}
    \begin{subfigure}{0.4\linewidth}
        \includegraphics[width=\textwidth]{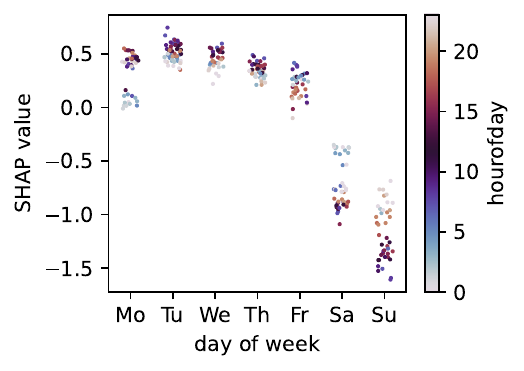}
    \end{subfigure}
    }
    \makebox[\linewidth][c]{
    \begin{subfigure}{0.4\linewidth}
        \includegraphics[width=\textwidth]{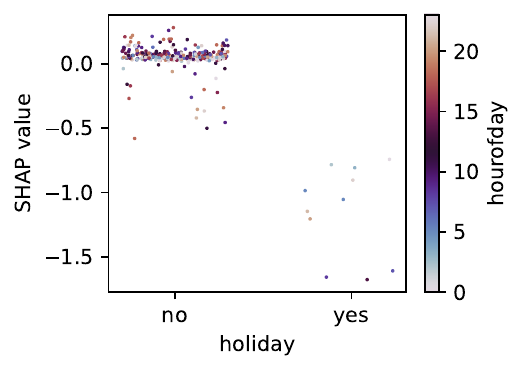}
    \end{subfigure}
    \begin{subfigure}{0.4\linewidth}
        \includegraphics[width=\textwidth]{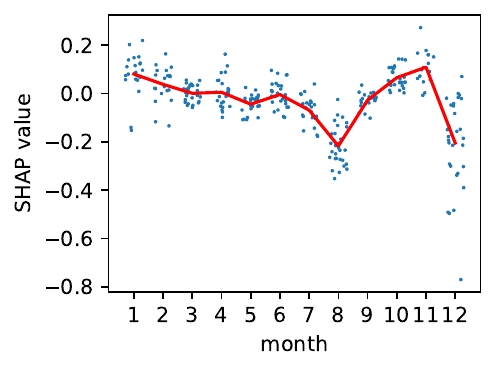}
    \end{subfigure}
    \begin{subfigure}{0.4\linewidth}
        \includegraphics[width=\textwidth]{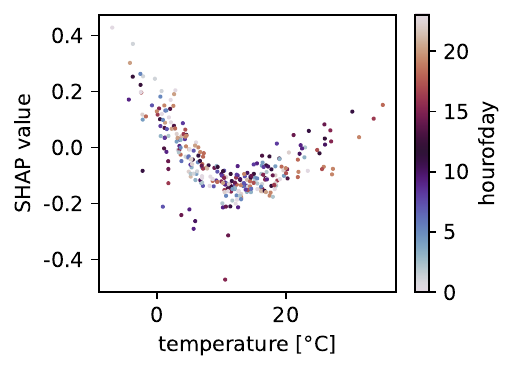}
    \end{subfigure}
    }

    \hspace{-1.5cm}B Custom Masker\\
    \makebox[\linewidth][c]{
    \begin{subfigure}{0.4\linewidth}
        \includegraphics[width=\textwidth]{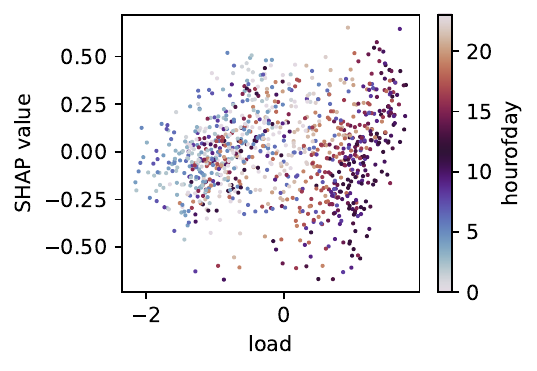}
    \end{subfigure}
    \begin{subfigure}{0.4\linewidth}
        \includegraphics[width=\textwidth]{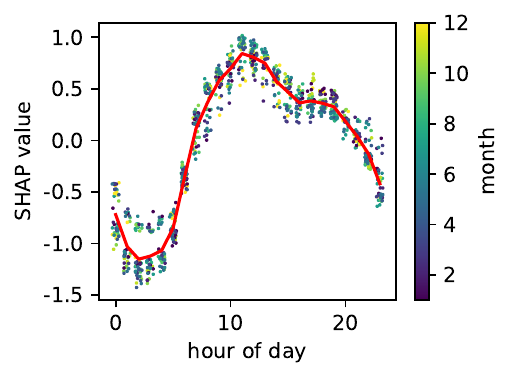}
    \end{subfigure}
    \begin{subfigure}{0.4\linewidth}
        \includegraphics[width=\textwidth]{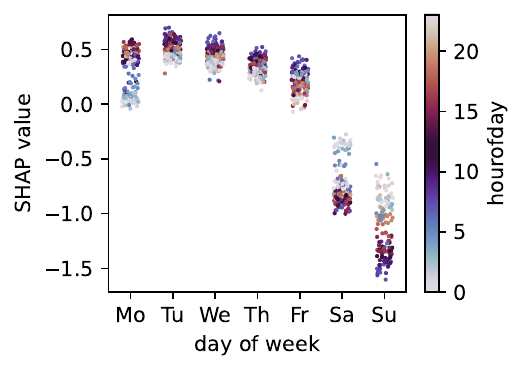}
    \end{subfigure}
    }
    \makebox[\linewidth][c]{
    \begin{subfigure}{0.4\linewidth}
        \includegraphics[width=\textwidth]{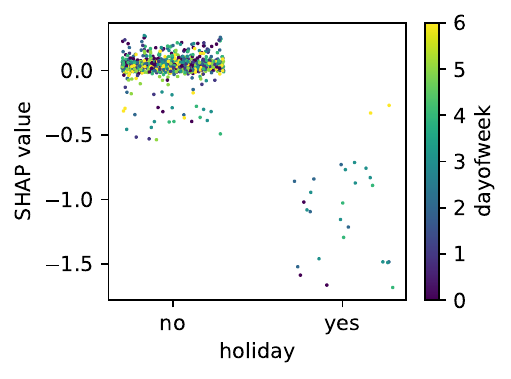}
    \end{subfigure}
    \begin{subfigure}{0.4\linewidth}
        \includegraphics[width=\textwidth]{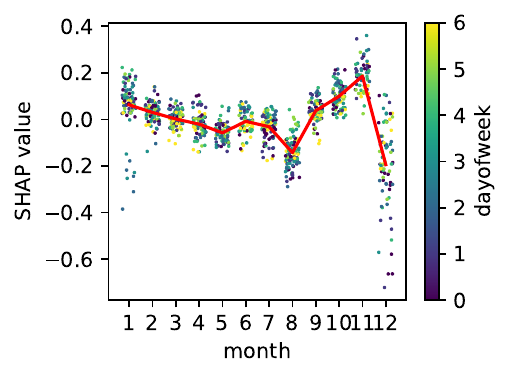}
    \end{subfigure}
    \begin{subfigure}{0.4\linewidth}
        \includegraphics[width=\textwidth]{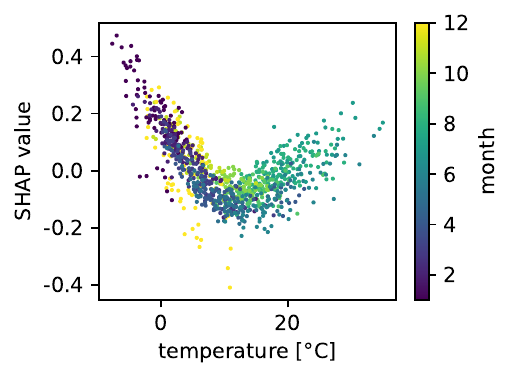}
    \end{subfigure}
    }
    
    \caption{Dependence plots on TransnetBW data created with the Permutation Explainer and the Custom Masker. For discrete variables, noise was added in the x-direction for visibility reasons.}
    \label{fig:global-explanations-transnet-shap}
\end{figure}

\begin{figure}
    \hspace{-1.5cm}A Data\\    
    \makebox[\linewidth][c]{
    \begin{subfigure}{0.8\linewidth}
        \includegraphics[width=\textwidth]{figures/legend_transnet.pdf}
    \end{subfigure}
    }
    \vspace{-1cm}
    
    \makebox[\linewidth][c]{
    \begin{subfigure}{0.6\linewidth}
        \includegraphics[scale=0.65]{figures/data/transnet/test_3684.pdf}
    \end{subfigure}
    \begin{subfigure}{0.6\linewidth}
        \includegraphics[scale=0.65]{figures/data/transnet/test_4077.pdf}
    \end{subfigure}
    }

    \hspace{-1.5cm}B Permutation Explainer\\
    \makebox[\linewidth][c]{
    \begin{subfigure}{0.6\linewidth}
        \includegraphics[width=\textwidth]{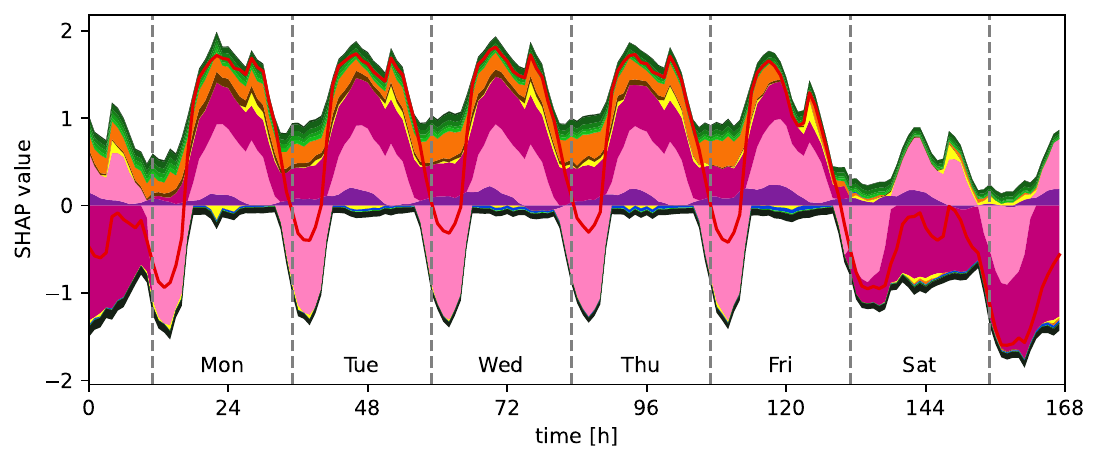}
    \end{subfigure}
    \begin{subfigure}{0.6\linewidth}
        \includegraphics[width=\textwidth]{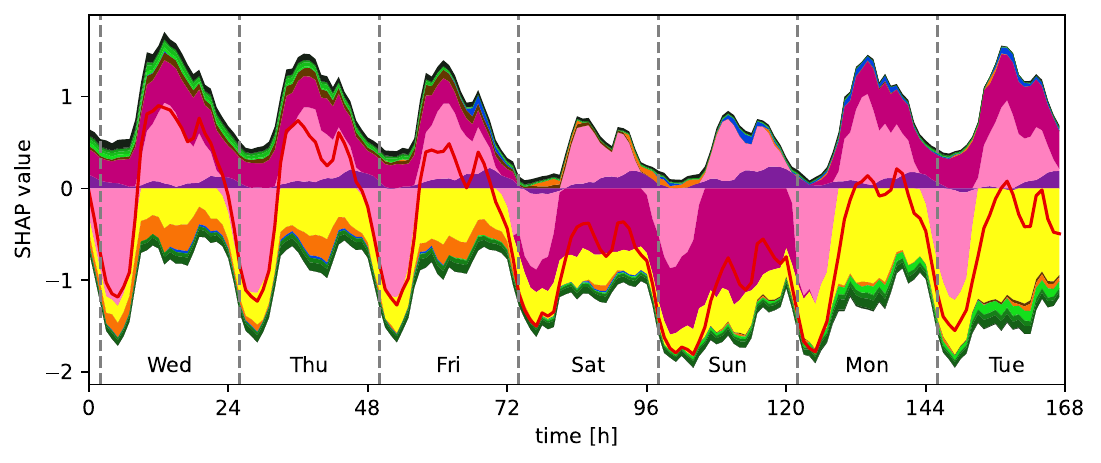}
    \end{subfigure}
    }

    \hspace{-1.5cm}C Custom Masker\\
    \makebox[\linewidth][c]{
    \begin{subfigure}{0.6\linewidth}
        \includegraphics[width=\textwidth]{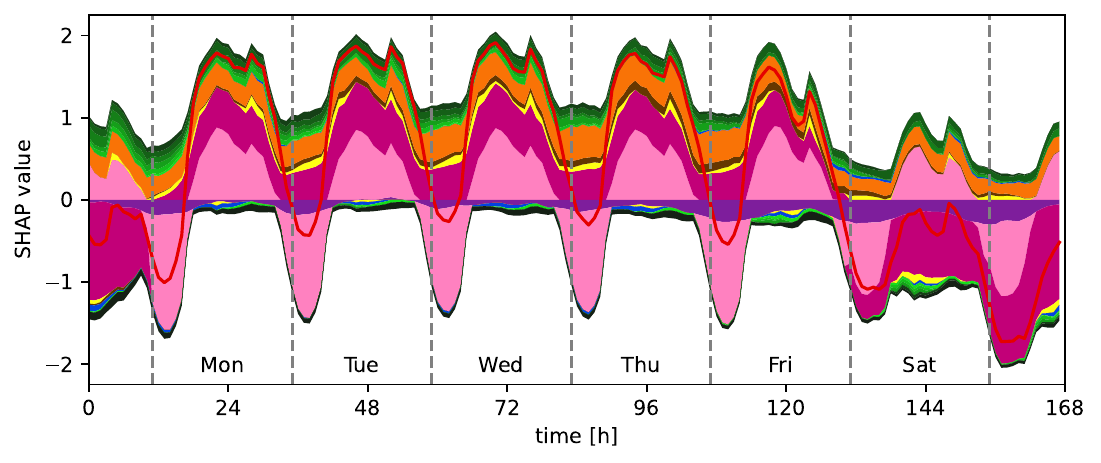}
    \end{subfigure}
    \begin{subfigure}{0.6\linewidth}
        \includegraphics[width=\textwidth]{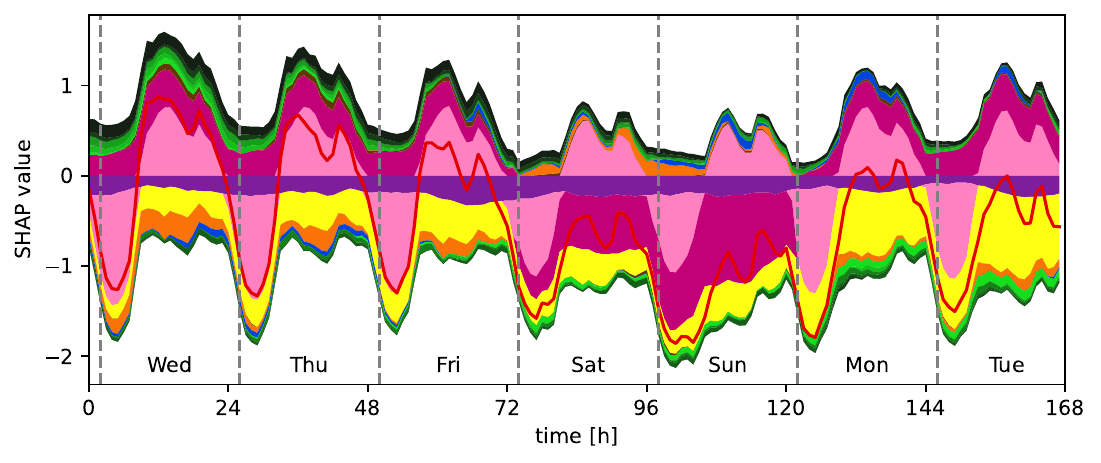}
    \end{subfigure}
    }
    
    \caption{Local explanations of TransnetBW examples created with the Permutation Explainer and the Custom Masker.}
    \label{fig:local-explanations-transnet-shap}
\end{figure}

See Figure \ref{fig:global-explanations-transnet-shap} for the global explanations and Figure \ref{fig:local-explanations-transnet-shap} for the local explanations on the TransnetBW data.

\end{appendices}


\end{document}